\documentclass{article} 
\usepackage{iclr2027_conference,times}

\usepackage{amsmath,amsfonts,bm}

\def\eqref#1{~(\ref{#1})}

\def\1{\bm{1}}

\DeclareMathAlphabet{\mathsfit}{\encodingdefault}{\sfdefault}{m}{sl}
\SetMathAlphabet{\mathsfit}{bold}{\encodingdefault}{\sfdefault}{bx}{n}

\usepackage{enumitem}
\usepackage{hyperref}
\usepackage{url}
\usepackage{graphicx}
\usepackage{svg}
\usepackage{algorithm}
\usepackage{algpseudocode}
\usepackage[table]{xcolor}
\usepackage{booktabs}
\usepackage{tabularx}
\usepackage{array}

\makeatletter
\@ifpackageloaded{booktabs}{}{\usepackage{booktabs}}
\@ifpackageloaded{multirow}{}{\usepackage{multirow}}
\@ifpackageloaded{fancyvrb}{}{\usepackage{fancyvrb}}
\@ifpackageloaded{tcolorbox}{}{\usepackage[most]{tcolorbox}}
\makeatother
\usepackage{fancyvrb}
\usepackage{fvextra}

\ifdefined\promptverb\else
\DefineVerbatimEnvironment{promptverb}{Verbatim}{%
  fontsize=\scriptsize,%
  frame=single,%
  framerule=0.4pt,%
  framesep=2mm,%
  breaklines=true,%
  breakautoindent=false,%
  baselinestretch=0.92}
\fi

\ifdefined\promptbox\else
\newtcolorbox{promptbox}[1]{%
  breakable, enhanced,%
  colback=gray!4, colframe=gray!55,%
  boxrule=0.5pt, arc=1pt,%
  left=2mm, right=2mm, top=1.5mm, bottom=1.5mm,%
  title={#1}, fonttitle=\bfseries\small,%
  coltitle=black, colbacktitle=gray!14}
\fi

\providecommand{\sid}[1]{\texttt{\detokenize{#1}}}

\title{
LLM Agents for Automated Post-Training via Structured Recipe Optimization
}

\author{Ruohong Liu\thanks{Equal contribution.}  \\
University of Oxford \\
\texttt{ruohong.liu@eng.ox.ac.uk}
\And
Bohan Tang\footnotemark[1] \\
LIGHTSPEED \\
\texttt{bohantang@global.tencent.com}
\And
Lekang Jiang \\
University of Cambridge \\
\texttt{lj408@cam.ac.uk}
\AND
Jack Umenberger\footnotemark[2] \\
University of Oxford \\
\texttt{jack.umenberger@eng.ox.ac.uk}
\And
Yiwen Guo\thanks{Corresponding author.} \\
Independent Researcher \\
\texttt{guoyiwen89@gmail.com}
}

\iclrfinalcopy 
\begin{document}

\maketitle
\lhead{Preprint}

\begin{abstract}
Automated post-training transforms pre-trained large language models (LLMs) into models capable of handling diverse domains by automatically constructing and optimizing a recipe of coupled decisions over data sources, data processing, cross-domain composition, and training configuration under a limited experimental budget. 
Existing automation either searches predefined recipe spaces that enable systematic optimization but remain fixed rather than adapting their available choices to the target tasks, or uses open-ended agents that discover task-specific choices but leave them implicit in trial-level edits, limiting their systematic reuse and recombination across trials.
To fill this gap, we introduce \textbf{optimization over a task-adapted recipe space}, a novel paradigm that can adapt the recipe space itself to the target tasks while representing its choices explicitly for systematic reuse, recombination, and joint optimization. We instantiate this paradigm with \textbf{Structured Recipe Optimization (SRO)}, comprising an LLM-based Recipe-Space Designer that constructs the task-adapted recipe space and a Multi-Domain Recipe Optimizer that uses structured trial evidence to guide LLM candidate construction and multi-objective Bayesian trial selection.
Extensive experiments show that SRO achieves the highest average score in all evaluated target-model and agent-LLM combinations. In joint post-training across three capability domains and six benchmarks, SRO reaches \textbf{41.1 versus 15.4} for the strongest baseline on Qwen3-1.7B with GPT-5.4, and \textbf{53.8 versus 40.8} on Qwen3-4B with DeepSeek-V4-Flash. Our code is available at \url{https://github.com/RuohLiuq/SRO}.

\end{abstract}

\section{Introduction}
Post-training transforms pre-trained large language models (LLMs) into instruction-following models capable of handling user requests across diverse domains, such as mathematical reasoning, coding, and knowledge-intensive question answering \citep{ouyang2022training,tie2025posttraining}. This process involves a sequence of coupled decisions over data processing, cross-domain composition, and training configuration, whose interactions can largely affect the resulting model \citep{tie2025posttraining,kumar2026llm}. Coordinating these decisions to form a post-training recipe requires substantial expert engineering \citep{grattafiori2024llama,guo2025deepseek}, while evaluating each candidate recipe requires actual training and multi-domain evaluation, making experimentation expensive. These considerations motivate automated post-training methods that automatically optimize for effective recipes under a limited experimental budget. 

Existing methods automate the overall post-training pipeline with two broad paradigms. First, \textit{optimization over a predefined recipe space} specifies a recipe space in advance
and systematically searches over the resulting alternatives for any given task.
For example, AutoPipe applies Bayesian optimization to post-training pipeline configurations \citep{chwa2026autopipe}.

While a predefined recipe space enables systematic comparison and
optimization, it may exclude useful task-specific choices, such as
data sources and processing methods, and thereby limit achievable
post-training performance.
Second, \textit{optimization over open-ended recipe space} instead delegates more of the design process
to LLM agents, allowing them to discover data, construct training
interventions, and revise recipes using evaluation feedback for each specific task
\citep{rank2026posttrainbench,yu2026autotrainess,chen2026agent2rlbench}. This enables greater task-specific adaptation, but newly discovered interventions are typically introduced within trial-level edits without being consolidated into explicit, reusable choices. Consequently, evidence from previous trials cannot be systematically reused to compare and recombine individual recipe decisions, making subsequent search less sample-efficient and reducing the likelihood of identifying high-performing recipe combinations under a limited training budget. Thereon, existing paradigms trade off \emph{task adaptability} against \emph{optimization effectiveness}. Predefined recipe spaces support effective optimization but can omit choices needed for the target task, while open-ended agents can use such choices but do not organize them into a space that can be efficiently optimized across trials. Both leave substantial post-training performance unrealized.

To fill this gap, we explore a novel paradigm termed \textbf{optimization over a task-adapted
recipe space}: constructing searchable choices for the target tasks
and jointly optimizing their combinations.
This paradigm introduces two key requirements. 
First, the recipe space must be \textit{task-adapted and structured}: it should incorporate task-relevant alternatives for data sources, data processing, cross-domain composition, and training configuration, while representing them as explicit, reusable, and composable choices whose combinations remain executable and comparable across trials. Second, optimization over this space must be \textit{joint and sample-efficient}. Since recipe components interact and each end-to-end evaluation is costly, search should reason over their combinations and use accumulated experimental evidence to prioritize promising recipes under a limited training budget.

\begin{figure*}[t]
    \centering
    \includegraphics[width=.85\textwidth]{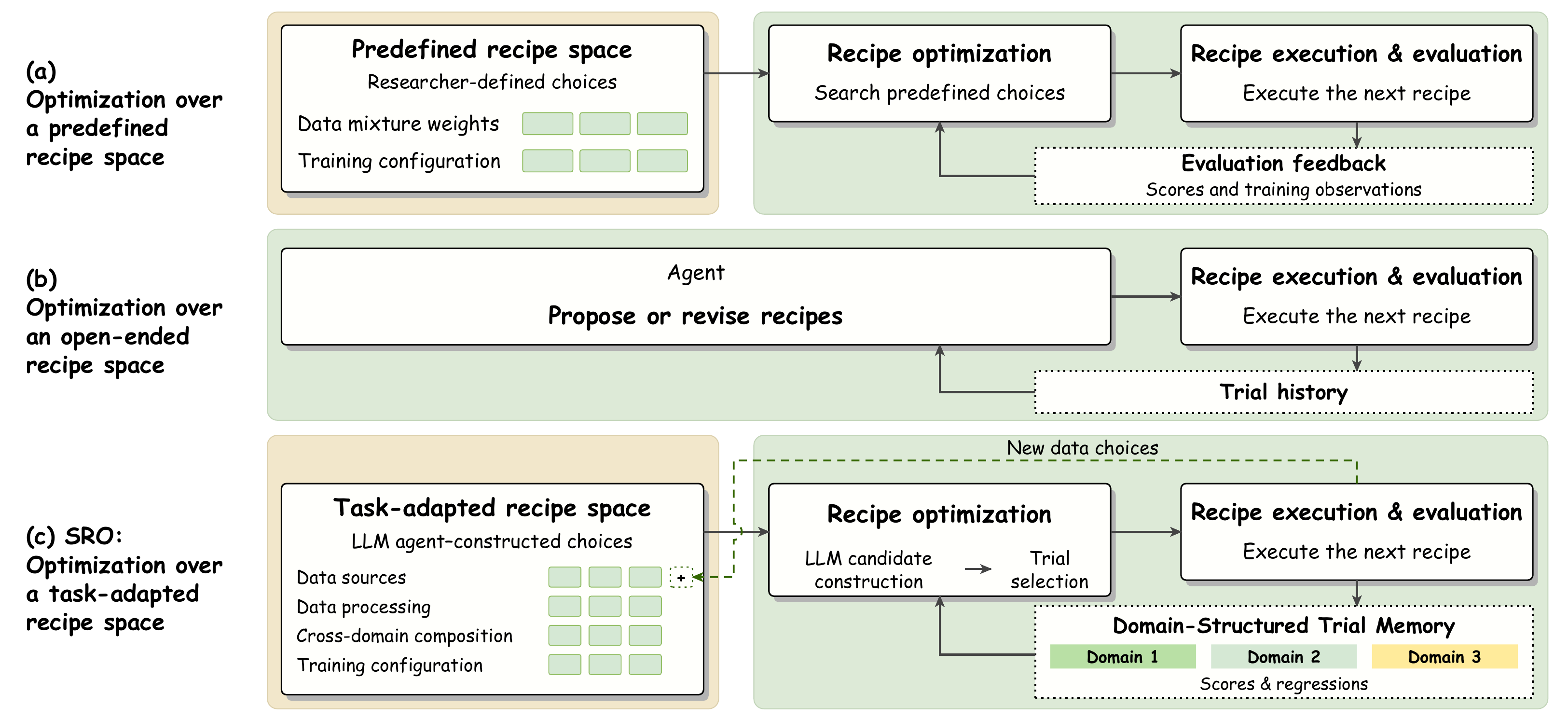}
    \caption{Three paradigms for automated post-training.
    }
    \label{fig:paradigm_comparison}
\end{figure*}

To realize this paradigm, we propose \textbf{Structured Recipe
Optimization (SRO)} with two key components.
A \textbf{Recipe-Space Designer} decomposes post-training into components for data sources, data processing, cross-domain composition, and training configuration, and asks the LLM agent to populate each component with choices based on the given task. We provide skills to guide this construction together with a fixed workflow for data preparation, training, evaluation, and validity checks, allowing newly designed choices to be incorporated without sacrificing executability or comparability.
A \textbf{Multi-Domain Recipe Optimizer} records executed recipes
and their benchmark scores, organizing the results by domain.
Selective retrieval provides relevant trial records and exploration
statistics to guide LLM construction of multiple candidate recipes
by recombining choices across components.
Multi-objective Bayesian optimization fits domain-wise surrogate
models to the trial history and uses Pareto-aware acquisition
to select one candidate for the next training trial. Figure~\ref{fig:paradigm_comparison} contrasts our proposed SRO with previous automated post-training paradigms.

Our main contributions are:

$\bullet$ We introduce \textbf{optimization over a task-adapted recipe
space} as a novel paradigm for automated post-training, representing
task-specific designs as explicit, reusable, and composable choices
to support joint search over data and training decisions.

$\bullet$ We propose \textbf{SRO} to instantiate this paradigm.
The Recipe-Space Designer constructs the search space, while the
Multi-Domain Recipe Optimizer organizes trial history to support
LLM candidate construction and multi-objective Bayesian trial
selection.

$\bullet$ Extensive experiments show that SRO achieves the highest average score in all evaluated
base-model and agent-LLM combinations.
For joint post-training across three capability domains and six
benchmarks, SRO reaches 41.1 versus 15.4 for the strongest baseline
on Qwen3-1.7B with GPT-5.4, and 53.8 versus 40.8 on Qwen3-4B with
DeepSeek-V4-Flash.

\section{Related Works}
\textbf{Post-Training.}
Post-training adapts a pre-trained language model to follow instructions and acquire desired capabilities across diverse downstream domains \citep{tie2025posttraining,kumar2026llm}. Modern post-training can be organized into three coupled stages: data preparation, supervised fine-tuning (SFT), and reinforcement learning (RL) \citep{tie2025posttraining,ouyang2022training,guo2025deepseek}. Data preparation constructs the supervision used by subsequent training stages by collecting, synthesizing, filtering, and mixing instruction--response pairs, reasoning trajectories, preference annotations, and task-specific feedback \citep{longpre2023flan,wang2023selfinstruct,cui2024ultrafeedback}. SFT trains the pre-trained model on curated samples to acquire instruction-following behavior and specific task capabilities \citep{wei2022finetuned,sanh2022multitask,longpre2023flan}. RL further optimizes the SFT model toward desired behaviors and capabilities through reward signals \citep{zhang2025survey}. For example, RL with human feedback (RLHF) improves alignment with human preferences, while RL with verifiable rewards (RLVR) strengthens reasoning ability by rewarding correct solutions and intermediate steps \citep{ouyang2022training,guo2025deepseek}.
LLM post-training involves designing a recipe over coupled decisions in data processing and training configurations, which demand substantial human expert engineering \citep{grattafiori2024llama,guo2025deepseek,tie2025posttraining,kumar2026llm}. Since each candidate recipe requires costly training and evaluation, automated methods are proposed to construct effective recipes automatically under limited experimental budgets.

\textbf{Automated Post-Training.} 
Existing work on automated post-training spans from component-level to pipeline-level automation. Component-level methods automate decisions within a fixed training procedure, such as data generation, selection, and mixture optimization for data preparation \citep{wang2023selfinstruct,xia2024less,zhang2026domainpilot}, sampling and resource allocation during SFT \citep{shin2026dynamixsft,gualdoni2026dynamics}, and domain balancing during RL \citep{li2025omni,yang2026tac,ramesh2026multitask}. More recent work moves toward pipeline-level automation, which is the focus of our work and broadly follows two paradigms. \textit{Optimization over a predefined recipe space} systematically searches over pipeline choices specified in advance. For instance, AutoPipe applies Bayesian optimization to a predefined configuration space with base-model choice and SFT/RL hyperparameters \citep{chwa2026autopipe}.
\textit{Optimization over an open-ended recipe space} instead delegates broader pipeline design to LLM agents. For instance, PostTrainBench gives agents full autonomy to discover training data and design and execute post-training strategies for target benchmarks \citep{rank2026posttrainbench}. However, existing approaches trade off task adaptability against optimization effectiveness: predefined recipe spaces enable systematic search but may omit task-specific choices, whereas open-ended agents can discover such choices but struggle to organize and optimize them across trials. We improve automated post-training performance by addressing this gap. 

\section{Methodology}

\subsection{Task Definition}
\label{sec:task_definition}
We study automated post-training over target domains
$\mathcal{D}=\{d_1,\ldots,d_K\}$, each evaluated by one or more
benchmarks.
The agent is given a pretrained base model $M_0$, the target domains
and benchmarks, a compute environment, and a time limit.
No training data are provided; the agent must identify or construct
suitable training data and specify the training configuration. Let $\mathcal{U}$ denote the set of executable post-training recipes permitted by these task constraints.
Executing a recipe $r\in\mathcal{U}$ produces a model
$M(r)=\operatorname{Train}(M_0;r)$.
Let $\mathbf{s}(M)$ denote the vector of benchmark scores obtained by model $M$, grouped by domain, and let $s_k(M)$ denote the mean score across the benchmarks in domain $d_k$.
The optimization objective is
\begin{equation}
    \max_{r\in\mathcal{U}} \bar{s}\bigl(M(r)\bigr),
    \qquad
    \bar{s}(M)=\frac{1}{K}\sum_{k=1}^{K}s_k(M).
    \label{eq:post_training_objective}
\end{equation}
At the end of the time limit, we select the checkpoint with the highest aggregate score during recipe optimization for final external evaluation. We solve this optimization through two coupled tasks.
First, \textbf{Recipe-Space Design} turns post-training decisions
such as data sources, data processing methods, cross-domain
composition, and training hyperparameters into searchable components
$\mathcal{C}=\{c_1,\ldots,c_J\}$.
Each component $c_j$ admits a set of choices
$\mathcal{V}_j=\{v_{j,1},\ldots,v_{j,n_j}\}$.
A recipe $r=(v_1,\ldots,v_J)$ selects one choice
$v_j\in\mathcal{V}_j$ for every component, and all valid, executable
combinations form the search space $\mathcal{R}\subseteq\mathcal{U}$.
Second, \textbf{Multi-Domain Recipe Optimization} searches this
space under the time limit.
At iteration $t$, the next recipe is determined from the current
search space $\mathcal{R}_{t-1}$ and experimental history
$\mathcal{H}_{t-1}$: $r_t=\operatorname{Optimizer}\bigl(\mathcal{R}_{t-1},\mathcal{H}_{t-1}\bigr)$
where $\operatorname{Optimizer}$ abstracts the feedback-driven search
procedure.
The selected recipe is executed and evaluated on all target
benchmarks, and the resulting scores and diagnostic evidence are
added to the history.

Predefined methods instantiate this process as
$r_t=\operatorname{Optimizer}
(\mathcal{R}_{\mathrm{fixed}},\mathcal{H}_{t-1})$,
using experimental feedback to guide search within a predefined
space $\mathcal{R}_{\mathrm{fixed}}\subseteq\mathcal{U}$.
Open-ended agentic methods instead operate over the permitted recipe
space, abstracted as
$r_t=\operatorname{Optimizer}(\mathcal{U},\mathcal{H}_{t-1})$.
They use experimental results, and potentially
structured memory to propose and revise recipes, without necessarily
consolidating the resulting choices into an explicit, reusable
search space.
We combine an agent-constructed recipe space with
domain-structured experimental evidence, enabling subsequent
decisions to draw on reusable choices and their observed
effects across domains.

\begin{figure*}[t]
    \centering
    \includegraphics[width=.9\textwidth]{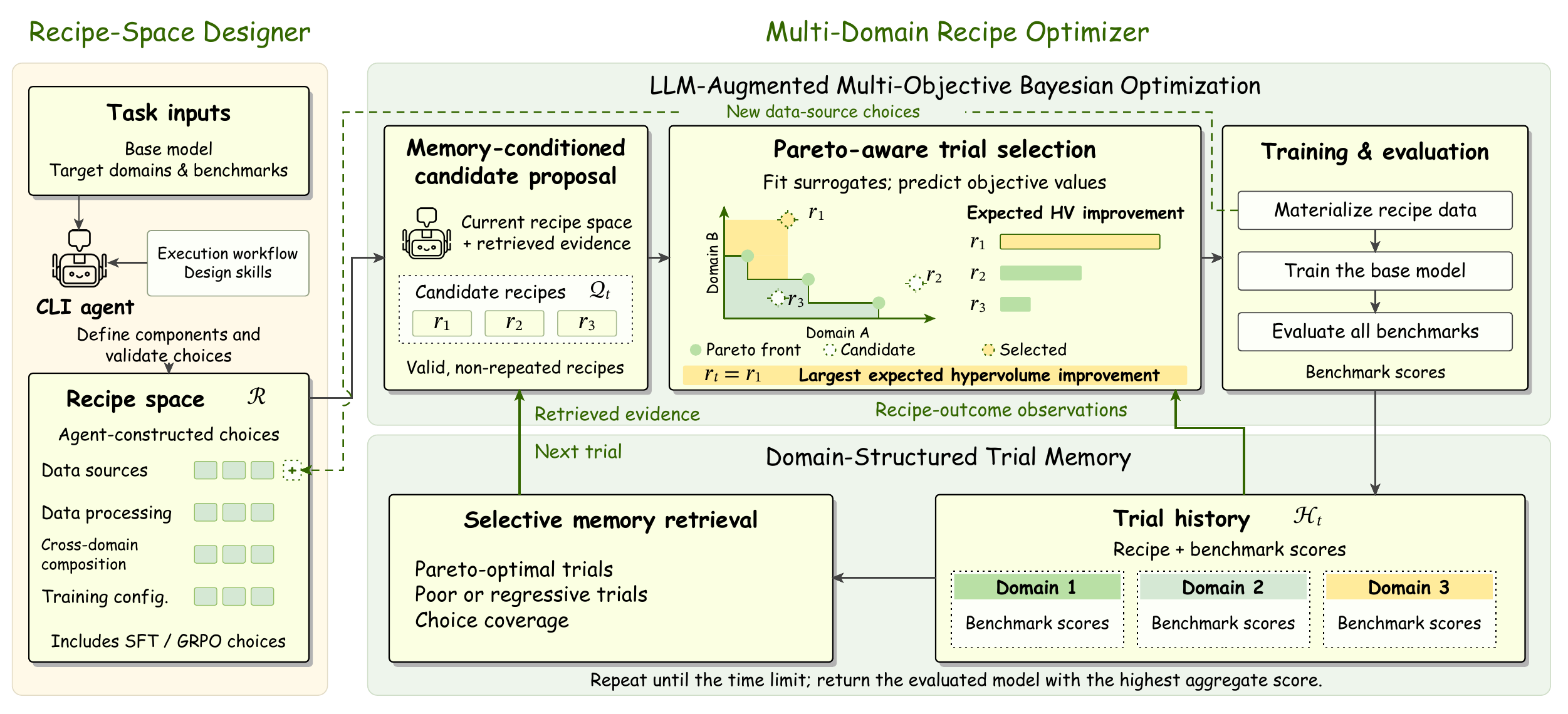}
\caption{
Overview of SRO.
The Recipe-Space Designer constructs a task-adapted recipe space, which the Multi-Domain Recipe Optimizer searches by Domain-Structured Trial Memory and LLM-Augmented Multi-Objective Bayesian Optimization.
}
    \label{fig:ampo_workflow}
\end{figure*}

\subsection{Structured Recipe Optimization (SRO)}
Structured Recipe Optimization (SRO) tackles the two coupled tasks in Section~\ref{sec:task_definition} with a Recipe-Space Designer and a Multi-Domain Recipe Optimizer. The Designer first constructs the initial task-adapted recipe space \(\mathcal R_0\). The Optimizer then iteratively uses trial history to propose and select a recipe, executes and evaluates it, and updates the trial history and recipe space for the next iteration. Figure~\ref{fig:ampo_workflow} summarizes this workflow.

\subsubsection{Recipe-Space Designer}
\label{sec:search-space}
The Recipe-Space Designer constructs the initial recipe space \(\mathcal R_0\) in three stages. It first defines the searchable recipe components, then provides an executable workflow and design skills for constructing valid choices, and finally instantiates and validates these choices to form \(\mathcal R_0\).

\paragraph{Searchable components.}
We factorize each recipe along four decision dimensions:
\textit{data sources}, \textit{data processing},
\textit{cross-domain composition}, and \textit{training configuration}.
The corresponding components form \(\mathcal{C} =
\left(\bigcup_{k=1}^{K}\mathcal{C}^{\mathrm{data}}_k\right)
\cup \mathcal{C}^{\mathrm{mix}}
\cup \mathcal{C}^{\mathrm{train}}\),
where $\mathcal{C}^{\mathrm{data}}_k$ contains the data-source and
data-processing components for domain $d_k$,
$\mathcal{C}^{\mathrm{mix}}$ contains the cross-domain composition, and $\mathcal{C}^{\mathrm{train}}$ contains the shared
training configuration.
This factorization defines the recipe-space schema, and the LLM agent subsequently instantiates task-adapted admissible choices for each component.

\paragraph{Execution workflow and design skills.}
Reimplementing data loading, formatting, training, and evaluation across trials wastes compute and introduces implementation variance.
We therefore provide a fixed executable workflow that performs these
operations, validates recipes, enforces compute constraints, and
excludes evaluation data from training.
To guide choice construction, we also provide reusable skills covering
dataset discovery, generation-prompt design, cross-domain composition,
and training configuration.
The workflow handles execution and validation, the skills provide design guidance, and the agent can thus focus on instantiating concrete recipe choices.

\paragraph{Choice construction and expansion.}
Using the component schema and design skills above, the LLM agent instantiates discrete choices for each searchable component.
For \textit{data sources}, the agent identifies datasets relevant to each target domain, inspects examples, and specifies the field mappings required by the executable workflow. Each validated dataset becomes a data-source choice.
For \textit{data processing}, we provide two data-generation templates for expanding capability coverage and addressing model deficiencies. For each instantiated template, the agent specifies a source dataset and generation prompt; generation runs only when selected by a recipe, strictly excludes evaluation data. Validated generated datasets are subsequently registered as new data-source choices.
For \textit{cross-domain composition}, the agent specifies mixture ratios and assembly modes under a fixed training-sample budget.
For \textit{training configuration}, the agent constructs configuration choices
for post-training algorithms such as SFT and GRPO.
The validated choices populate the component-wise sets $\{\mathcal{V}_j\}_{j=1}^{J}$, whose valid combinations define the initial recipe space $\mathcal{R}_0$.

\subsubsection{Multi-Domain Recipe Optimizer}
\label{sec:recipe-optimizer}
The Multi-Domain Recipe Optimizer searches the current recipe space using feedback from previous trials.
At iteration $t$, Domain-Structured Trial Memory retrieves a compact summary $\mathcal{S}_{t-1}$ from the trial history $\mathcal{H}_{t-1}$ and current recipe space $\mathcal{R}_{t-1}$.
An LLM proposer uses this summary to construct candidate recipes $\mathcal{Q}_t$, and multi-objective Bayesian optimization selects the next recipe $r_t$ using the complete trial history.
The selected recipe is then executed and evaluated, updating the trial history and, when new data-source choices are produced, the recipe space for the next iteration.

\paragraph{Domain-Structured Trial Memory.} This memory retains each completed trial as a structured record
$\xi_i=(r_i,\mathbf{s}(M_i))$, containing the executed recipe and its scores on all target domains.
The scores are organized by domain to preserve differences in cross-domain performance, and the history before iteration $t$ is denoted by $\mathcal{H}_{t-1}$. The history supports optimization in two paths.
The Bayesian optimizer uses the history to fit its surrogate models, while the LLM proposer receives a compact summary obtained by selective retrieval: $\mathcal{S}_{t-1}
=\operatorname{Retrieve}(\mathcal{H}_{t-1},\mathcal{R}_{t-1})$.
The summary includes Pareto-optimal or otherwise strong trials, lower-performing trials, aggregate statistics over explored choices, and coverage statistics identifying under-explored choices.
Retaining recipe configurations alongside domain-wise results allows the proposer to compare historical recipes and their effects across domains rather than relying only on aggregate scores.

\paragraph{LLM-Augmented Multi-Objective Bayesian Optimization.}
Recipe optimization separates \emph{candidate construction} from
\emph{trial selection}.
Since proposing recipes with an LLM agent is inexpensive relative to data preparation, model training, and evaluation, we use the agent to construct multiple candidates and Bayesian optimization to select which candidate to execute.
At iteration $t$, the LLM receives the current recipe space
$\mathcal{R}_{t-1}$ and retrieved summary $\mathcal{S}_{t-1}$, producing
$\mathcal{Q}_t=G_{\mathrm{LLM}}(\mathcal{R}_{t-1},\mathcal{S}_{t-1})$.
It fills structured recipe fields with admissible choices rather than
implementing a new training pipeline.
The proposer compares recipe configurations and their domain-wise
results across retrieved trials, together with choice-coverage statistics.
These comparisons guide the reuse of promising configurations,
the recombination of existing choices, and the exploration of
under-explored options within the current recipe space.

The Bayesian optimizer uses the complete history $\mathcal{H}_{t-1}$
to predict each candidate's domain-wise outcomes and uncertainty.
The mean benchmark score within each domain defines an objective,
modeled by a separate surrogate using executed recipes and their
observed outcomes.
A Pareto-aware acquisition function combines the predictive distributions
to identify candidates expected to improve the current cross-domain
trade-offs.
We use expected hypervolume improvement (EHVI) and select
$r_t=\arg\max_{r\in\mathcal{Q}_t}\alpha_t(r)$,
where $\alpha_t(r)$ is the acquisition score.
Intuitively, EHVI measures the expected increase in dominated hypervolume
relative to a reference point, accounting for predictive uncertainty.
A warm-up strategy is used when too few completed trials are available
to fit the surrogates.
Appendix~\ref{app:bo-details} describes surrogate fitting and acquisition
computation.

\paragraph{Trial execution and update.} The executable workflow executes the selected recipe $r_t$, performing data preparation, training, and evaluation to obtain $M_t$ and its benchmark scores.
Data generation uses the sources and prompts from the
Recipe-Space Designer.
Correction uses the latest trained model to identify failed
examples in its separate source dataset and generate corrective
training data.
While these failed examples are used to prompt data generation, the examples themselves are not retained in trial memory.
After evaluation, the completed trial is added to the history $\mathcal{H}_{t}=\mathcal{H}_{t-1}\cup\{\xi_{t}\}$.
Validated generated datasets are registered as new data-source choices, producing the updated recipe space $\mathcal{R}_t$.
The updated states are the inputs to the next optimization iteration.

\section{Experiments}
We evaluate SRO through three questions concerning overall performance,
explicit recipe-space optimization, and the mechanisms used to search
the constructed space.

\begin{center}
\setlength{\fboxsep}{8pt}
\colorbox{gray!8}{%
\begin{minipage}{0.94\linewidth}
\textbf{Research Questions}
\vspace{2pt}

\begin{enumerate}[leftmargin=18pt]
    \setlength{\itemsep}{3pt}
    \setlength{\parskip}{0pt}

    \item \textbf{RQ1:}
    How does SRO compare with existing automated post-training methods?

    \item \textbf{RQ2:}
    How does our Recipe-Space Designer support recipe optimization?

    \item \textbf{RQ3:}
    How does our Multi-Domain Recipe Optimizer support recipe optimization?
\end{enumerate}
\end{minipage}%
}
\end{center}

\subsection{Experimental Setup}
\label{experiment}
\paragraph{Tasks and evaluation.}
We evaluate SRO in single-domain and multi-domain settings using the
benchmarks and evaluation protocols provided by
PostTrainBench~\citep{rank2026posttrainbench}.
Each single-domain run targets one of GSM8K, GPQA Main, or HumanEval. Each multi-domain run jointly targets six benchmarks across three domains: mathematical reasoning (AIME 2025 and GSM8K), coding and tool use (BFCL v3 \texttt{exec\_simple} and HumanEval), and science and health (GPQA Main and HealthBench-Easy). The multi-domain aggregate score is the mean of the three domain scores, each computed by averaging its two benchmarks. All scores are reported as percentages.

\paragraph{Models and agent LLMs.} We post-train Qwen3-1.7B-Base and Qwen3-4B-Base using DeepSeek-V4-Flash and GPT-5.4 as agent LLMs through the OpenCode CLI. Within each base-model and agent-LLM configuration, all methods use the same agent LLM and CLI scaffold. We additionally report the official instruction-tuned models as reference points rather than competing methods because their post-training budgets are not matched.

\paragraph{Baselines.}
We compare against PostTrainBench~\citep{rank2026posttrainbench} and
AutoTrainess~\citep{yu2026autotrainess}, two end-to-end agent-based
methods that automate data acquisition, training, and evaluation.
These methods align with our focus on the agent's ability to automate
the complete post-training pipeline without requiring any predefined
training data.
PostTrainBench allows the agent to autonomously design and revise
its training approach, while AutoTrainess provides a structured workflow.
Both baselines run through OpenCode with the same base model,
task specification, evaluation interface, agent LLM, and compute
budget as SRO.
We additionally evaluate \emph{optimization over a predefined recipe
space} on GSM8K, whose official training split provides a
task-specific training source.
Appendix~\ref{app:predefined-space} describes this supplementary
baseline and its results.

We also evaluate four variants with DeepSeek-V4-Flash:
\textit{w/o structured recipe search} replaces explicit recipe-space construction and its associated executable workflow with direct open-ended agentic recipe design;
\textit{w/o LLM proposer} uniformly samples candidate recipes from the
same recipe space without using LLM proposals;
\textit{w/o EHVI selection} omits surrogate fitting and EHVI, selecting
LLM-proposed candidates by self-reported confidence;
and \textit{w/o structured memory} replaces selective retrieval with
the full history of recipes and raw benchmark scores, without
domain-structured summaries or choice-coverage statistics.
In the last variant, the history used for surrogate fitting is unchanged.

\paragraph{Training and compute budget.}
All runs start from the provided base model. The training-configuration space includes LoRA-based SFT and, for single-domain runs, an optional GRPO phase. Each run uses one GPU with a wall-clock budget of 10 hours for
single-domain experiments and 20 hours for multi-domain experiments.
All methods follow the evaluation protocols and data-contamination
constraints provided by PostTrainBench. We select the checkpoint with the highest aggregate score during recipe optimization
and report its full-set evaluation outside the time budget
(Appendix~\ref{app:setup:benchmarks}).

\subsection{RQ1: Overall Effectiveness}
\label{sec:overall-effectiveness}

\paragraph{SRO achieves the highest average score in both settings.}
Tables~\ref{tab:main_singlebench_results}
and~\ref{tab:main_sixbench_results} show that SRO achieves
the highest average score in every evaluated base-model
and agent-LLM configuration in both settings.
In single-domain post-training, SRO improves individual
benchmark scores by up to 41.7 points over the strongest
baseline in the same configuration
(GSM8K, Qwen3-4B with GPT-5.4: 87.4\% vs.\ 45.7\%).
In multi-domain post-training, SRO improves the joint
six-benchmark average by up to 25.7 points
(Qwen3-1.7B with GPT-5.4: 41.1\% vs.\ 15.4\%).

\begin{table*}[ht]
\centering
\caption{
Single-domain post-training with two base models and two agent LLMs.
Each run targets one benchmark; Avg. denotes the mean score across the
three separate runs. 
}
\label{tab:main_singlebench_results}
\resizebox{.9\textwidth}{!}{
\begin{tabular}{lllcccc}
\toprule
& & &
\multicolumn{3}{c}{\textbf{Benchmark Scores} ($\uparrow$)} & \\
\cmidrule(lr){4-6}
\textbf{Model} &
\textbf{Agent LLM} &
\textbf{Harness} &
\textbf{GSM8K} &
\textbf{GPQA Main} &
\textbf{HumanEval} &
\textbf{Avg.} ($\uparrow$) \\
\midrule

\multirow{7}{*}{Qwen3-1.7B-Base}
& -- & None
& 12.7 & 14.1 & 7.9 & 11.6 \\
\cmidrule(lr){2-7}

& \multirow{3}{*}{DeepSeek-V4-Flash}
& PostTrainBench
& 49.0 & 27.5 & 54.9 & 43.8 \\
& & AutoTrainess
& \textbf{79.1} & 28.1 & 53.7 & 53.6 \\
& &
\cellcolor{gray!12}\textbf{SRO (Ours)}
& \cellcolor{gray!12}68.2
& \cellcolor{gray!12}\textbf{34.2}
& \cellcolor{gray!12}\textbf{64.6}
& \cellcolor{gray!12}\textbf{55.6} \\
\cmidrule(lr){2-7}

& \multirow{3}{*}{GPT-5.4}
& PostTrainBench
& 48.4 & 28.6 & 39.0 & 38.7 \\
& & AutoTrainess
& 61.4 & 14.3 & 16.5 & 30.7 \\
& &
\cellcolor{gray!12}\textbf{SRO (Ours)}
& \cellcolor{gray!12}\textbf{69.5}
& \cellcolor{gray!12}\textbf{32.4}
& \cellcolor{gray!12}\textbf{53.7}
& \cellcolor{gray!12}\textbf{51.9} \\

\cmidrule(lr){1-7}
Qwen3-1.7B-Instruct
& -- & Reference
& 88.5 & 35.5 & 68.9 & 64.3 \\

\midrule

\multirow{7}{*}{Qwen3-4B-Base}
& -- & None
& 41.8 & 13.4 & 36.6 & 30.6 \\
\cmidrule(lr){2-7}

& \multirow{3}{*}{DeepSeek-V4-Flash}
& PostTrainBench
& 77.3 & 33.9 & 67.7 & 59.6 \\
& & AutoTrainess
& 71.2 & 24.3 & 75.6 & 57.0 \\
& &
\cellcolor{gray!12}\textbf{SRO (Ours)}
& \cellcolor{gray!12}\textbf{87.6}
& \cellcolor{gray!12}\textbf{35.5}
& \cellcolor{gray!12}\textbf{77.4}
& \cellcolor{gray!12}\textbf{66.8} \\
\cmidrule(lr){2-7}

& \multirow{3}{*}{GPT-5.4}
& PostTrainBench
& 45.7 & 24.1 & 73.2 & 47.7 \\
& & AutoTrainess
& 42.8 & 19.0 & 39.6 & 33.8 \\
& &
\cellcolor{gray!12}\textbf{SRO (Ours)}
& \cellcolor{gray!12}\textbf{87.4}
& \cellcolor{gray!12}\textbf{35.0}
& \cellcolor{gray!12}\textbf{77.4}
& \cellcolor{gray!12}\textbf{66.6} \\

\cmidrule(lr){1-7}
Qwen3-4B-Instruct
& -- & Reference
& 93.8 & 44.6 & 77.4 & 71.9 \\

\bottomrule
\end{tabular}
}
\end{table*}

\begin{table*}[ht]
\centering
\caption{
Multi-domain post-training across six benchmarks in three domains.
Each run jointly targets all six benchmarks. Avg. is the mean across
domains.
}
\label{tab:main_sixbench_results}
\resizebox{.95\textwidth}{!}{
\begin{tabular}{lllccccccc}
\toprule
& & &
\multicolumn{6}{c}{\textbf{Benchmark Scores} ($\uparrow$)} & \\
\cmidrule(lr){4-9}
& & &
\multicolumn{2}{c}{\textbf{Mathematical Reasoning}} &
\multicolumn{2}{c}{\textbf{Coding \& Tool Use}} &
\multicolumn{2}{c}{\textbf{Science \& Health}} & \\
\cmidrule(lr){4-5}
\cmidrule(lr){6-7}
\cmidrule(lr){8-9}
\textbf{Model} &
\textbf{Agent LLM} &
\textbf{Harness} &
\textbf{AIME 2025} &
\textbf{GSM8K} &
\textbf{BFCL} &
\textbf{HumanEval} &
\textbf{GPQA Main} &
\textbf{HealthBench} &
\textbf{Avg.} ($\uparrow$) \\
\midrule

\multirow{7}{*}{Qwen3-1.7B-Base}
& -- & None
& 0.0 & 12.7 & 0.0 & 7.9 & 14.1 & 7.5 & 7.0 \\
\cmidrule(lr){2-10}

& \multirow{3}{*}{DeepSeek-V4-Flash}
& PostTrainBench
& 0.0 & 58.8 & \textbf{92.0} & 35.4 & 27.9 & 8.6 & 37.1 \\
& & AutoTrainess
& 0.0 & 73.8 & \textbf{92.0} & \textbf{58.5} & \textbf{30.8} & 15.6 & 45.1 \\
& &
\cellcolor{gray!12}\textbf{SRO (Ours)}
& \cellcolor{gray!12}\textbf{6.7}
& \cellcolor{gray!12}\textbf{76.7}
& \cellcolor{gray!12}91.0
& \cellcolor{gray!12}\textbf{58.5}
& \cellcolor{gray!12}16.1
& \cellcolor{gray!12}\textbf{22.7}
& \cellcolor{gray!12}\textbf{45.3} \\
\cmidrule(lr){2-10}

& \multirow{3}{*}{GPT-5.4}
& PostTrainBench
& 0.0 & 40.0 & 0.0 & 0.0 & 7.4 & 7.2 & 9.1 \\
& & AutoTrainess
& 0.0 & 13.2 & 0.0 & 42.7 & 23.9 & 12.4 & 15.4 \\
& &
\cellcolor{gray!12}\textbf{SRO (Ours)}
& \cellcolor{gray!12}\textbf{3.3}
& \cellcolor{gray!12}\textbf{57.1}
& \cellcolor{gray!12}\textbf{93.0}
& \cellcolor{gray!12}\textbf{45.7}
& \cellcolor{gray!12}\textbf{32.6}
& \cellcolor{gray!12}\textbf{15.2}
& \cellcolor{gray!12}\textbf{41.1} \\

\cmidrule(lr){1-10}
Qwen3-1.7B-Instruct
& -- & Reference
& 26.7 & 88.5 & 94.0 & 68.9 & 35.5 & 44.9 & 59.7 \\

\midrule

\multirow{7}{*}{Qwen3-4B-Base}
& -- & None
& 3.3 & 41.8 & 0.0 & 36.6 & 13.4 & 13.4 & 18.1 \\
\cmidrule(lr){2-10}

& \multirow{3}{*}{DeepSeek-V4-Flash}
& PostTrainBench
& 0.0 & 64.1 & 88.0 & 45.7 & \textbf{32.1} & 14.7 & 40.8 \\
& & AutoTrainess
& 0.0 & 55.6 & 88.0 & 43.9 & 26.8 & 15.8 & 38.4 \\
& &
\cellcolor{gray!12}\textbf{SRO (Ours)}
& \cellcolor{gray!12}\textbf{3.3}
& \cellcolor{gray!12}\textbf{87.6}
& \cellcolor{gray!12}\textbf{94.0}
& \cellcolor{gray!12}\textbf{79.9}
& \cellcolor{gray!12}30.8
& \cellcolor{gray!12}\textbf{26.9}
& \cellcolor{gray!12}\textbf{53.8} \\
\cmidrule(lr){2-10}

& \multirow{3}{*}{GPT-5.4}
& PostTrainBench
& 0.0 & 43.4 & 0.0 & 36.6 & 15.0 & 10.4 & 17.6 \\
& & AutoTrainess
& 0.0 & 49.7 & \textbf{90.0} & 48.2 & 28.8 & 13.4 & 38.4 \\
& &
\cellcolor{gray!12}\textbf{SRO (Ours)}
& \cellcolor{gray!12}0.0
& \cellcolor{gray!12}\textbf{85.7}
& \cellcolor{gray!12}88.0
& \cellcolor{gray!12}\textbf{59.8}
& \cellcolor{gray!12}\textbf{29.0}
& \cellcolor{gray!12}\textbf{22.6}
& \cellcolor{gray!12}\textbf{47.5} \\

\cmidrule(lr){1-10}
Qwen3-4B-Instruct
& -- & Reference
& 53.3 & 93.8 & 95.0 & 77.4 & 44.6 & 52.7 & 69.5 \\

\bottomrule
\end{tabular}
}
\end{table*}

\paragraph{SRO shows lower average sensitivity to the agent LLM.}
SRO matches or exceeds the strongest baseline in 11 of the 12
single-domain model--agent--benchmark combinations. The exception is
GSM8K on Qwen3-1.7B with DeepSeek-V4-Flash, where AutoTrainess reaches
79.1 versus 68.2. In multi-domain experiments on Qwen3-4B with
DeepSeek-V4-Flash, SRO improves all three domain averages over the
strongest baseline by aggregate score: 45.5 versus 32.1 in mathematical
reasoning, 87.0 versus 66.9 in coding and tool use, and 28.9 versus
23.4 in science and health. SRO also exhibits lower average sensitivity
to the agent LLM in multi-domain experiments. Averaged across the two
base models, switching agents changes its aggregate score by 5.3 points,
compared with 25.6 for PostTrainBench and 14.9 for AutoTrainess.

\subsection{RQ2: Recipe-Space Designer Effectiveness}
\label{sec:recipe-space-results}

\paragraph{A task-adapted and structured recipe space supports effective
post-training.}
The \textit{w/o structured recipe search} variant retains structured trial memory
but does not construct a task-adapted recipe space before search.
Instead, the agent directly designs and executes recipes during
experimentation.
Under the same compute budget, its average score falls from 45.3
to 34.5 on Qwen3-1.7B and from 53.8 to 40.3 on Qwen3-4B
(Table~\ref{tab:ablation_results}).
These results support constructing a task-adapted, structured recipe
space and using the associated workflow to search its combinations.

\begin{table*}[t]
\centering
\caption{
Comparison with ablations of the recipe space design and search mechanisms.
All runs use DeepSeek-V4-Flash and the same compute budget on each base
model. 
}
\label{tab:ablation_results}
\resizebox{.9\textwidth}{!}{
\begin{tabular}{llccccccc}
\toprule
& & \multicolumn{6}{c}{\textbf{Benchmark Scores} ($\uparrow$)} & \\
\cmidrule(lr){3-8}
& & \multicolumn{2}{c}{\textbf{Mathematical Reasoning}} &
\multicolumn{2}{c}{\textbf{Coding \& Tool Use}} &
\multicolumn{2}{c}{\textbf{Science \& Health}} & \\
\cmidrule(lr){3-4}
\cmidrule(lr){5-6}
\cmidrule(lr){7-8}
\textbf{Model} &
\textbf{Variant} &
\textbf{AIME 2025} &
\textbf{GSM8K} &
\textbf{BFCL} &
\textbf{HumanEval} &
\textbf{GPQA Main} &
\textbf{HealthBench} &
\textbf{Avg.} ($\uparrow$) \\
\midrule

\rowcolor{gray!12}
\multirow{5}{*}{Qwen3-1.7B}
& \textbf{Full \textsc{SRO}}
& \textbf{6.7} & \textbf{76.7} & 91.0 & 58.5 & 16.1 & \textbf{22.7} & \textbf{45.3} \\
& w/o structured recipe search
& 0.0 & 57.3 & 85.0 & 34.1 & 23.2 & 7.6 & 34.5 \\
& w/o LLM proposer
& \textbf{6.7} & 74.3 & \textbf{94.0} & 55.5 & 20.3 & 13.4 & 44.0 \\
& w/o EHVI selection
& 0.0 & 56.0 & \textbf{94.0} & 51.2 & \textbf{29.7} & 15.4 & 41.1 \\
& w/o structured memory
& 3.3 & 75.7 & 91.0 & \textbf{59.8} & 16.5 & 13.8 & 43.4 \\
\midrule

\rowcolor{gray!12}
\multirow{5}{*}{Qwen3-4B}
& \textbf{Full \textsc{SRO}}
& 3.3 & 87.6 & 94.0 & \textbf{79.9} & 30.8 & \textbf{26.9} & \textbf{53.8} \\
& w/o structured recipe search
& 6.7 & 63.1 & 90.0 & 45.1 & 30.4 & 6.7 & 40.3 \\
& w/o LLM proposer
& 3.3 & 86.6 & \textbf{95.0} & 72.6 & 22.3 & 21.6 & 50.2 \\
& w/o EHVI selection
& \textbf{10.0} & \textbf{88.4} & 93.0 & 69.5 & \textbf{36.4} & 14.2 & 51.9 \\
& w/o structured memory
& 0.0 & 81.3 & 93.0 & 65.9 & 33.9 & 15.3 & 48.2 \\
\bottomrule
\end{tabular}
}
\end{table*}

\begin{figure*}[t]
    \centering
    \includegraphics[width=.95\textwidth]{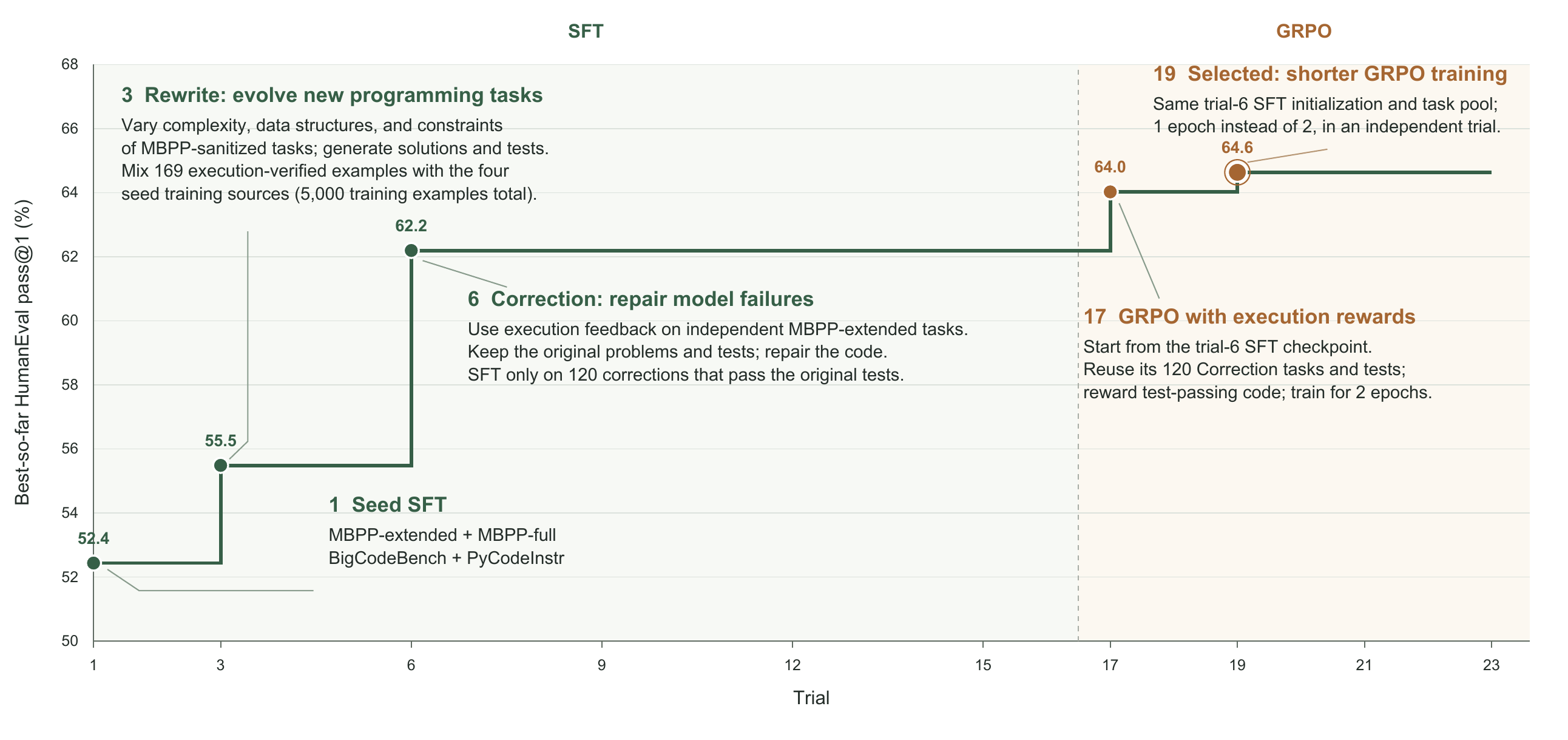}
    \caption{
    Search trajectory on HumanEval for Qwen3-1.7B-Base with
    DeepSeek-V4-Flash. Each point denotes an evaluated recipe, and the
    solid step curve tracks the best HumanEval pass@1 observed so far.
    The vertical dashed line separates the SFT and GRPO phases.
    Numbered annotations mark the seed SFT recipe, the introduction of
    Rewrite-generated data, the addition of Correction data, the first
    GRPO improvement, and the final selected checkpoint.
    }
    \label{fig:singlebench_search_trajectory}
\end{figure*}

\paragraph{Expanding the recipe space with reusable choices improves subsequent recipes.}
Figure~\ref{fig:singlebench_search_trajectory} shows the search
on HumanEval with Qwen3-1.7B and DeepSeek-V4-Flash.
The agent initially identifies MBPP and MBPP+ as data-source choices
and specifies sources and generation prompts for Rewrite and
Correction. These choices, together with training-configuration
choices, form the initial recipe space.
During search, selected data-processing methods generate training
data, and validated datasets become additional data-source choices
for subsequent recipes.
The initial SFT recipe achieves 52.4, while subsequent recipes using
Rewrite and Correction data reach 55.5 and 62.2, respectively.
Correction uses model errors on a separate source dataset, not
HumanEval evaluation examples.
GRPO initialized from the best SFT model then reaches 64.0;
adjusting training configurations yields 64.6.
This shows how search incorporates newly generated data
as reusable choices and combines them with training choices to
improve subsequent recipes.

\subsection{RQ3: Multi-Domain Recipe Optimizer Effectiveness}
\label{sec:rq3_ablation}

\begin{figure*}[t]
    \centering
    \includegraphics[width=.95\textwidth]{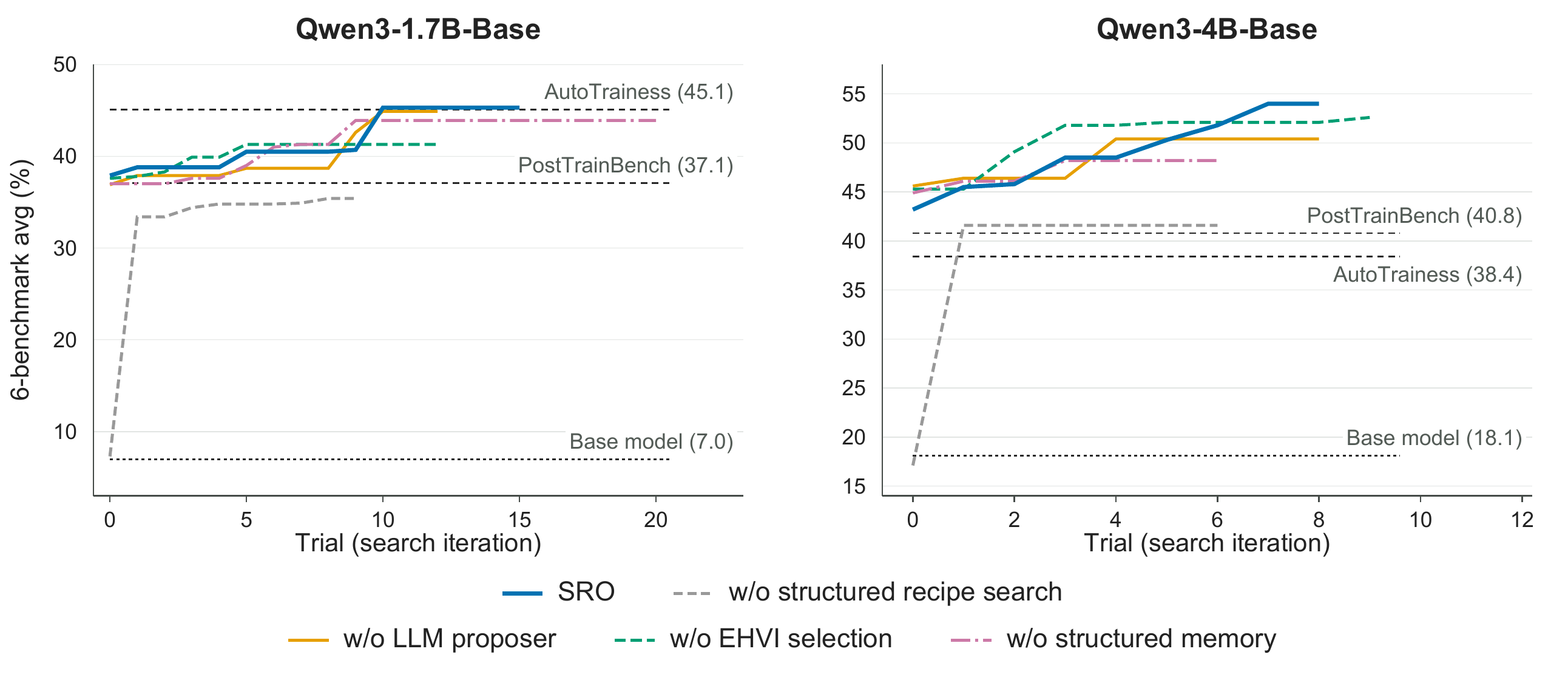}
    \caption{
    Optimization trajectories of SRO and its variants under the same
    compute budget. Each curve tracks the best multi-domain average
    obtained over completed trials.
    }
    \label{fig:ablation_trajectories}
\end{figure*}

We next examine candidate proposal, trial selection, and structured
memory within the designed recipe space, using the remaining ablations
in Table~\ref{tab:ablation_results} and
Figure~\ref{fig:ablation_trajectories}.

\paragraph{LLM candidate construction and EHVI trial selection both
improve performance.}
Replacing the LLM proposer with uniform candidate sampling reduces
the final score by 1.3 points on Qwen3-1.7B and 3.6 points on Qwen3-4B.
The resulting scores of 44.0 and 50.2 show that the constructed space
and retained EHVI selection still support effective search, while
LLM proposals provide further gains. Replacing EHVI with LLM-reported
confidence reduces the scores by 4.2 and 1.9 points, respectively.
On Qwen3-4B, confidence-based selection improves several individual
benchmarks but lowers HealthBench-Easy from 26.9 to 14.2, yielding a
lower aggregate score. These results support using surrogate-based
selection rather than the proposer's confidence to compare candidates
with different cross-domain effects.

\paragraph{Selective retrieval and structured summaries improve
candidate construction.}
Giving the proposer the full history of recipes and raw benchmark
scores, without selective retrieval, domain-structured summaries, or
choice-coverage statistics, reduces the final score by 1.9 points on
Qwen3-1.7B and 5.6 points on Qwen3-4B.
Both SRO and this variant retain the executed recipes and their benchmark scores,
and the history used for surrogate fitting is unchanged. The comparison
therefore supports organizing and selectively retrieving historical
results for the proposer, rather than supplying the complete raw history.

\section{Conclusion}
\label{sec:conclusion}

We introduce \textbf{optimization over a task-adapted recipe space}, a novel paradigm for automated post-training that adapt the recipe space to the target tasks while representing its choices explicitly for systematic reuse, recombination, and joint optimization. Thereon, we propose \textbf{Structured Recipe Optimization (SRO)}, which constructs a task-adapted recipe space and uses accumulated multi-domain trial evidence to guide sample-efficient recipe optimization. Experiments show that SRO consistently outperforms existing automated post-training methods, offering an approach that combines task adaptability with systematic optimization under limited experimental budgets.


\bibliography{iclr2027_conference}

@misc{chen2026agent2rlbench,
title={Agent$^2$ RL-Bench: Can LLM Agents Engineer Agentic RL Post-Training?}, 
author={Wanyi Chen and Xiao Yang and Xu Yang and Tianming Sha and Qizheng Li and Zhuo Wang and Bowen Xian and Fang Kong and Weiqing Liu and Jiang Bian},
year={2026},
eprint={2604.10547},
archivePrefix={arXiv},
primaryClass={cs.AI},
url={https://arxiv.org/abs/2604.10547}, 
}

@article{ouyang2022training,
  title={Training language models to follow instructions with human feedback},
  author={Ouyang, Long and Wu, Jeffrey and Jiang, Xu and Almeida, Diogo and Wainwright, Carroll and Mishkin, Pamela and Zhang, Chong and Agarwal, Sandhini and Slama, Katarina and Ray, Alex and others},
  journal={Advances in neural information processing systems},
  volume={35},
  pages={27730--27744},
  year={2022}
}

@article{grattafiori2024llama,
  title={The llama 3 herd of models},
  author={Grattafiori, Aaron and Dubey, Abhimanyu and Jauhri, Abhinav and Pandey, Abhinav and Kadian, Abhishek and Al-Dahle, Ahmad and Letman, Aiesha and Mathur, Akhil and Schelten, Alan and Vaughan, Alex and others},
  journal={arXiv preprint arXiv:2407.21783},
  year={2024},
  url           = {https://arxiv.org/abs/2407.21783}
}

@article{guo2025deepseek,
  title={Deepseek-r1: Incentivizing reasoning capability in llms via reinforcement learning},
  author={Guo, Daya and Yang, Dejian and Zhang, Haowei and Song, Junxiao and Wang, Peiyi and Zhu, Qihao and Xu, Runxin and Zhang, Ruoyu and Ma, Shirong and Bi, Xiao and others},
  journal={arXiv preprint arXiv:2501.12948},
  year={2025}
}

@article{xia2024less,
  title={Less: Selecting influential data for targeted instruction tuning},
  author={Xia, Mengzhou and Malladi, Sadhika and Gururangan, Suchin and Arora, Sanjeev and Chen, Danqi},
  journal={arXiv preprint arXiv:2402.04333},
  year={2024}
}

@article{rank2026posttrainbench,
  title={PostTrainBench: Can LLM Agents Automate LLM Post-Training?},
  author={Rank, Ben and Bhatnagar, Hardik and Prabhu, Ameya and Eisenberg, Shira and Nguyen, Karina and Bethge, Matthias and Andriushchenko, Maksym},
  journal={arXiv preprint arXiv:2603.08640},
  year={2026}
}

@inproceedings{wei2022finetuned,
  title     = {Finetuned Language Models Are Zero-Shot Learners},
  author    = {Wei, Jason and Bosma, Maarten and Zhao, Vincent Y. and Guu, Kelvin and Yu, Adams Wei and Lester, Brian and Du, Nan and Dai, Andrew M. and Le, Quoc V.},
  booktitle = {International Conference on Learning Representations},
  year      = {2022},
  url       = {https://arxiv.org/abs/2109.01652}
}

@inproceedings{sanh2022multitask,
  title     = {Multitask Prompted Training Enables Zero-Shot Task Generalization},
  author    = {Sanh, Victor and Webson, Albert and Raffel, Colin and Bach, Stephen H. and Sutawika, Lintang and Alyafeai, Zaid and Chaffin, Antoine and Stiegler, Arnaud and Scao, Teven Le and Raja, Arun and Dey, Manan and Bari, M Saiful and Xu, Canwen and Thakker, Urmish and Sharma, Shanya Sirohi and Szczechla, Eliza and Kim, Taewoon and Chhablani, Gunjan and Nayak, Nihal and Datta, Debajyoti and Chang, Jonathan and Jiang, Mike Tian-Jian and Wang, Han and Manica, Matteo and Shen, Sheng and Yong, Zheng Xin and Pandey, Harshit and Bawden, Rachel and Wang, Thomas and Neeraj, Trishala and Rozen, Jos and Sharma, Abheesht and Santilli, Andrea and Fevry, Thibault and Fries, Jason Alan and Teehan, Ryan and Biderman, Stella and Gao, Leo and Bers, Thomas and Wolf, Thomas and Rush, Alexander M.},
  booktitle = {International Conference on Learning Representations},
  year      = {2022},
  url       = {https://arxiv.org/abs/2110.08207}
}

@InProceedings{longpre2023flan,
  title = 	 {The Flan Collection: Designing Data and Methods for Effective Instruction Tuning},
  author =       {Longpre, Shayne and Hou, Le and Vu, Tu and Webson, Albert and Chung, Hyung Won and Tay, Yi and Zhou, Denny and Le, Quoc V and Zoph, Barret and Wei, Jason and Roberts, Adam},
  booktitle = 	 {Proceedings of the 40th International Conference on Machine Learning},
  pages = 	 {22631--22648},
  year = 	 {2023},
  editor = 	 {Krause, Andreas and Brunskill, Emma and Cho, Kyunghyun and Engelhardt, Barbara and Sabato, Sivan and Scarlett, Jonathan},
  volume = 	 {202},
  series = 	 {Proceedings of Machine Learning Research},
  month = 	 {23--29 Jul},
  publisher =    {PMLR},
  url = 	 {https://proceedings.mlr.press/v202/longpre23a.html},
}

@inproceedings{wang2023selfinstruct,
  title     = {Self-Instruct: Aligning Language Models with Self-Generated Instructions},
  author    = {Wang, Yizhong and Kordi, Yeganeh and Mishra, Swaroop and Liu, Alisa and Smith, Noah A. and Khashabi, Daniel and Hajishirzi, Hannaneh},
  booktitle = {Proceedings of the 61st Annual Meeting of the Association for Computational Linguistics (Volume 1: Long Papers)},
  pages     = {13484--13508},
  year      = {2023},
  month     = jul,
  doi       = {10.18653/v1/2023.acl-long.754},
  url       = {https://aclanthology.org/2023.acl-long.754/}
}

@inproceedings{
cui2024ultrafeedback,
title={{ULTRAFEEDBACK}: Boosting Language Models with Scaled {AI} Feedback},
author={Ganqu Cui and Lifan Yuan and Ning Ding and Guanming Yao and Bingxiang He and Qiang Yue and Yuan Ni and Guotong Xie and Ruobing Xie and Yankai Lin and Zhiyuan Liu and Maosong Sun},
booktitle={Forty-first International Conference on Machine Learning},
year={2024},
url={https://openreview.net/forum?id=BOorDpKHiJ}
}

@inproceedings{shin2026dynamixsft,
  title     = {{DynamixSFT}: Dynamic Mixture Optimization of Instruction Tuning Collections},
  author    = {Shin, Haebin and Ji, Lei and Liu, Xiao and Yu, Zhiwei and Yoo, Hyunwoo and Chen, Qi and Gong, Yeyun},
  booktitle = {Findings of the Association for Computational Linguistics: ACL 2026},
  pages     = {39590--39603},
  year      = {2026},
  month     = jul,
  doi       = {10.18653/v1/2026.findings-acl.1972},
  url       = {https://aclanthology.org/2026.findings-acl.1972/}
}

@article{gualdoni2026dynamics,
  title         = {{DynaMiCS}: Fine-Tuning {LLMs} with Performance Constraints Using Dynamic Mixtures},
  author        = {Gualdoni, Eleonora and Laguna, Sonia and Bethune, Louis and Monteiro, Joao and Ablin, Pierre and Cuturi, Marco},
  journal       = {arXiv preprint arXiv:2605.10770},
  year          = {2026},
  eprint        = {2605.10770},
  archivePrefix = {arXiv},
  url           = {https://arxiv.org/abs/2605.10770}
}

@article{zhang2026domainpilot,
  title         = {{DomainPilot}: Domain-Level Loss-Guided Two-Stage Data Mixture Optimization for Efficient Language Model Fine-Tuning},
  author        = {Zhang, He},
  journal       = {arXiv preprint arXiv:2607.22769},
  year          = {2026},
  eprint        = {2607.22769},
  archivePrefix = {arXiv},
  url           = {https://arxiv.org/abs/2607.22769}
}

@article{yang2026tac,
  title         = {Transferability for General Reasoning: An Automated Curriculum for Multi-Domain {RLVR}},
  author        = {Yang, Yongjin and Liu, Jiarui and He, Yinghui and Zhang, Lechen and Sch{\"o}lkopf, Bernhard and Jin, Zhijing},
  journal       = {arXiv preprint arXiv:2606.25178},
  year          = {2026},
  eprint        = {2606.25178},
  archivePrefix = {arXiv},
  url           = {https://arxiv.org/abs/2606.25178}
}

@article{chwa2026autopipe,
  title         = {Automatic Configuration of {LLM} Post-Training Pipelines},
  author        = {Chwa, Channe and Wu, Xinle and Lu, Yao},
  journal       = {arXiv preprint arXiv:2603.18773},
  year          = {2026},
  eprint        = {2603.18773},
  archivePrefix = {arXiv},
  url           = {https://arxiv.org/abs/2603.18773}
}

@article{tie2025posttraining,
  title={A survey on post-training of large language models},
  author={Tie, Guiyao and Zhao, Zeli and Song, Dingjie and Wei, Fuyang and Zhou, Rong and Dai, Yurou and Yin, Wen and Yang, Zhejian and Yan, Jiangyue and Su, Yao and others},
  journal={arXiv preprint arXiv:2503.06072},
  year={2025},
  url           = {https://arxiv.org/abs/2503.06072}
}

@article{zhang2025survey,
  title={A survey of reinforcement learning for large reasoning models},
  author={Zhang, Kaiyan and Zuo, Yuxin and He, Bingxiang and Sun, Youbang and Liu, Runze and Jiang, Che and Fan, Yuchen and Tian, Kai and Jia, Guoli and Li, Pengfei and others},
  journal={arXiv preprint arXiv:2509.08827},
  year={2025},
  url           = {https://arxiv.org/abs/2509.08827}
}

@article{kumar2026llm,
  title={Llm post-training: A deep dive into reasoning large language models},
  author={Kumar, Komal and Ashraf, Tajamul and Thawakar, Omkar and Anwer, Rao Muhammad and Cholakkal, Hisham and Shah, Mubarak and Yang, Ming-Hsuan and Torr, Phillip HS and Khan, Fahad Shahbaz and Khan, Salman},
  journal={IEEE Transactions on Pattern Analysis and Machine Intelligence},
  year={2026},
  publisher={IEEE},
  url           = {https://ieeexplore.ieee.org/abstract/document/11640952}
}

@article{li2025omni,
  title={Omni-thinker: Scaling multi-task rl in llms with hybrid reward and task scheduling},
  author={Li, Derek and Zhou, Jiaming and Brunswic, Leo Maxime and Ghaddar, Abbas and Sun, Qianyi and Ma, Liheng and Luo, Yu and Li, Dong and Coates, Mark and Hao, Jianye and others},
  journal={arXiv preprint arXiv:2507.14783},
  year={2025},
  url           = {https://arxiv.org/abs/2507.14783}
}

@inproceedings{
ramesh2026multitask,
title={Multi-Task {GRPO}: Reliable {LLM} Reasoning Across Tasks},
author={Shyam Sundhar Ramesh and Xiaotong Ji and Matthieu Zimmer and Sangwoong Yoon and Zhiyong Wang and Haitham Bou Ammar and Aurelien Lucchi and Ilija Bogunovic},
booktitle={Forty-third International Conference on Machine Learning},
year={2026},
url={https://openreview.net/forum?id=zA3KhRZ9A5}
}

@article{yu2026autotrainess,
  title={AutoTrainess: Teaching Language Models to Improve Language Models Autonomously},
  author={Yu, Zhaojian and Yin, Penghao and Gao, Shuzheng and He, Shilin and Cai, Kai and Zhang, Xiao-Ping},
  journal={arXiv preprint arXiv:2606.31551},
  year={2026}
}

@article{jones1998efficient,
  title={Efficient global optimization of expensive black-box functions},
  author={Jones, Donald R and Schonlau, Matthias and Welch, William J},
  journal={Journal of Global optimization},
  volume={13},
  number={4},
  pages={455--492},
  year={1998},
  publisher={Springer}
}

@inproceedings{hutter2011sequential,
  title={Sequential model-based optimization for general algorithm configuration},
  author={Hutter, Frank and Hoos, Holger H and Leyton-Brown, Kevin},
  booktitle={International conference on learning and intelligent optimization},
  pages={507--523},
  year={2011},
  organization={Springer}
}

@article{emmerich2006single,
  title={Single-and multiobjective evolutionary optimization assisted by Gaussian random field metamodels},
  author={Emmerich, Michael TM and Giannakoglou, Kyriakos C and Naujoks, Boris},
  journal={IEEE Transactions on Evolutionary Computation},
  volume={10},
  number={4},
  pages={421--439},
  year={2006},
  publisher={IEEE}
}
\bibliographystyle{iclr2027_conference}

\newpage
\appendix
\section{Limitations}

\textbf{Dynamic recipe-space evolution.}
SRO constructs a task-adapted recipe space before optimization and can expand it during search when newly generated datasets become reusable choices within existing components. However, the set of searchable components and their organization remain fixed within each run: optimization does not introduce new component types or modify the component hierarchy. Building on SRO's structured recipe representation, future work could extend this flexibility from choice-level expansion to schema-level evolution, allowing new components or hierarchical structures to be introduced as experimental evidence accumulates.

\textbf{Continual post-training across tasks and models.}
SRO uses accumulated trial evidence to guide recipe optimization for a fixed set of target tasks and a given base model. While this enables reuse across trials within a run, we do not study a continual setting in which new post-training tasks or base models are encountered sequentially and experience from earlier runs is carried forward. Future work could extend SRO to reuse recipe choices, trial evidence, or optimization priors across successive post-training problems, reducing repeated exploration and enabling increasingly efficient automated post-training over time.

\section{Experimental Setup Details}
\label{app:setup}

\subsection{Benchmarks, Evaluation Protocol, and Scoring}
\label{app:setup:benchmarks}

We consider single-domain and multi-domain post-training.
Single-domain runs target GSM8K, GPQA Main, or HumanEval separately.
Multi-domain runs jointly target six benchmarks spanning mathematical
reasoning, coding and tool use, and science and health.
Table~\ref{tab:app:benchmarks} summarizes the evaluation sets,
output formats, and scoring rules.
Within each setting, all compared methods use the same evaluation
scripts, prompts, and chat templates provided by
PostTrainBench~\citep{rank2026posttrainbench}.
The agent may not modify these evaluation components.
HealthBench-Easy uses GPT-5.4-mini as the judge, with the same judge
configuration across all compared methods.

\begin{table}[h]
\centering
\caption{Evaluation benchmarks, output formats, and scoring rules
for the single-domain and multi-domain experiments.}
\label{tab:app:benchmarks}
\small
\begin{tabular}{lllll}
\toprule
Benchmark & Domain & Test size & Output contract & Scoring \\
\midrule
AIME 2025 & math & 30 & \sid{ANSWER: <integer>} & exact match \\
GSM8K & math & 1{,}319 & \sid{ANSWER: <number>} & exact match \\
BFCL v3 & coding \& tool use & 500 & structured \sid{tool_calls} & schema execution \\
HumanEval & coding \& tool use & 164 & complete Python function & unit tests \\
GPQA Main & science \& health & 448 & \sid{ANSWER: <A--D>} & exact match \\
HealthBench-Easy & science \& health & 245  & free-form responses & LLM-judge rubrics \\
\bottomrule
\end{tabular}
\end{table}

During SRO recipe optimization, each trial is evaluated on 50\% of the examples from each target benchmark, except HealthBench-Easy, for which 50 samples are used. These scores guide recipe optimization and checkpoint selection. For PostTrainBench and AutoTrainess, the agents choose their search-time evaluation sample sizes through the provided evaluation interface.

At the end of each run, the checkpoint with the highest observed score during SRO recipe optimization is selected and evaluated on the complete target benchmark sets outside the time budget. For PostTrainBench and AutoTrainess, the agent select from the models derived from iterative optimization which it decides as final model based on evaluation on the target benchmark.
For single-domain runs, we report the target benchmark score; for multi-domain runs, we report the mean across domains, each computed by averaging its benchmark scores.

\subsection{Compute Budget and Infrastructure}
\label{app:setup:compute}

Each multi-domain run (SRO and both baselines) receives
a 20-hour wall-clock budget that covers discovery, data generation, training,
and search-time evaluation. Each single-domain run
(Table~\ref{tab:main_singlebench_results}) receives a 10-hour budget.
Final independent evaluation is conducted outside this budget.
All methods receive the same budget, task description, evaluation interface,
and hardware. Each run has exclusive access to one GPU with
16{,}896 CUDA cores, 132 streaming multiprocessors, and 141\,GB of memory.

Every method---SRO, the
PostTrainBench baseline, and the AutoTrainess baseline---drives the same CLI
agent scaffold (OpenCode) with one of two agent LLMs, DeepSeek-V4-Flash or
GPT-5.4, served through a fixed gateway. In SRO the same agent LLM serves as
(i) the discovery agent that constructs the recipe space, (ii) the in-loop
recipe proposer, and (iii) the framework's auxiliary data-generation calls (Rewrite /
Correction workers).

A SRO run consists
of (i) a \emph{recipe space design phase} in which the CLI agent explores sources and
methods and submits the executable handoff, and (ii) a \emph{recipe optimization phase}
in which it executes at most iterative rounds of proposal, acquisition,
training, and evaluation, until reaching
wall-clock budget.

\subsection{Training Configuration}
\label{app:setup:training}

SRO runs start from the provided base model, which is never replaced or
merged with external model weights. 

Warmup ratio is fixed at 0.05.
Qwen3-1.7B-Base uses per-device batch size 8 with gradient accumulation 4;
Qwen3-4B-Base uses batch size 2 with accumulation 16 (effective batch size 32
in both cases). The SFT loss is applied to the assistant target only.

In single-domain runs, after three consecutive
completed SFT trials with neither realized Pareto-hypervolume growth
nor any improvement in domain-wise historical-best scores, the search
enters a GRPO phase initialized from the best SFT checkpoint and
continues LoRA training. Training prompts come from discovered sources
with reference answers or executable tests. Rewards use the verification
rules of the target benchmark to check answer correctness or unit-test
success; evaluation examples never enter the training prompt pool.
None of the reported multi-domain runs entered this phase within the
time budget; all results in Table~\ref{tab:main_sixbench_results}
therefore come from SFT trials.

For GRPO, the proposer searches five components: learning rate
$\in \{5{\times}10^{-7}, 10^{-6}, 2{\times}10^{-6}\}$,
KL coefficient $\beta \in \{0, 0.02, 0.04\}$, group size
$G \in \{8, 16\}$, training budget (\sid{short} or \sid{long}),
and prompt selector (\sid{all} or \sid{unstable}).
The \sid{short} and \sid{long} budgets correspond to one and two
passes over the prompt pool, respectively. The \sid{unstable} selector
retains only prompts for which the current model's sampled answers
disagree; \sid{all} uses the complete pool.
The remaining settings are fixed. Rollout temperature is 0.8 and
top-$p$ is 0.95. Maximum completion length is 512 tokens, and maximum
prompt length is half the model's sequence-length limit. Per-device
batch size is half the SFT value, with gradient accumulation adjusted
so that the effective batch size is divisible by $G$. Training uses
bf16 precision.

\subsection{Bayesian Optimizer Implementation}
\label{app:bo-details}

For Bayesian optimization, benchmark scores are normalized using
the base and instruction-tuned reference scores in
Table~\ref{tab:app:references}, then averaged within each domain
to define the optimization objectives.

\begin{table}[h]
\centering
\caption{Base and instruction-tuned reference scores used to normalize
the BO objectives. Scores are reported on a percentage scale.}
\label{tab:app:references}
\small
\begin{tabular}{lccccccc}
\toprule
Model & AIME & GSM8K & BFCL & HumanEval & GPQA & HealthBench & Mean \\
\midrule
Qwen3-1.7B-Base       & 0.0  & 12.7 & 0.0  & 7.9  & 14.1 & 7.5  & 7.0  \\
Qwen3-1.7B-Instruct   & 26.7 & 88.5 & 94.0 & 68.9 & 35.5 & 44.9 & 59.7 \\
Qwen3-4B-Base         & 3.3  & 41.8 & 0.0  & 36.6 & 13.4 & 13.4 & 18.1 \\
Qwen3-4B-Instruct     & 53.3 & 93.8 & 95.0 & 77.4 & 44.6 & 52.7 & 69.5 \\
\bottomrule
\end{tabular}
\end{table}

Whenever Bayesian selection is used, its hyperparameters are fixed as
listed in Table~\ref{tab:app:bo-hyper}. Each recipe $r$ is encoded as
a feature vector $\phi(r)$ consisting of the per-domain budget shares, a
one-hot encoding of the assembly mode, one binary indicator per registered
data source (including sources generated during the search), one binary
indicator per data-generation method, and the training choices for the
active phase. The SFT and GRPO choices are specified in
Appendix~\ref{app:setup:training}; the feature dimension depends on
the recipe space and training phase.

For each parent domain, an independent random forest regresses $\phi(r)$
onto the corresponding normalized domain objective. Each forest consists of 24 regression
trees of maximum depth 4, each trained on a bootstrap replica of the trial
history over a random 70\% subset of the features. We use random forests
to model the predominantly categorical recipe features
\citep{hutter2011sequential}. The trees' mean yields the prediction,
and their disagreement provides an empirical uncertainty estimate.

For each candidate recipe $r$, we
generate $S=64$ outcome vectors by independently sampling one tree's
prediction from each domain's forest. We estimate expected hypervolume
improvement (EHVI) \citep{emmerich2006single} as
\begin{equation}
\alpha_t(r)
=
\frac{1}{S}\sum_{m=1}^{S}
\left[
\operatorname{HV}\!\left(
\mathcal{P}_{t-1}\cup\{\mathbf{y}^{(m)}(r)\}
\right)
-\operatorname{HV}(\mathcal{P}_{t-1})
\right],
\label{eq:mc-ehvi}
\end{equation}
where $\mathcal{P}_{t-1}$ is the Pareto front of observed normalized
domain scores and $\mathbf{y}^{(m)}(r)$ is the $m$-th sampled outcome
vector. Here, $\operatorname{HV}$ measures dominated hypervolume
relative to the normalized base-model reference point. Each round
executes the candidate with the highest acquisition score. In
single-domain runs, the objective is scalar and the acquisition
estimates improvement over the best observed score
\citep{jones1998efficient}.

Until five feasible evaluated trials are available, we use a warm-up
heuristic instead of fitting the forests. It is based on the
proposing LLM's self-declared priors,
$c_{\mathrm{effect}} + 0.4\,\mathrm{conf} + 0.2\,\mathrm{novelty}
- 0.3\,\mathrm{risk}$; this rule is active only during the earliest
rounds. Cross-domain draws treat the per-domain predictions as independent,
so cross-domain error correlations are not modeled. Selection is
sequential, with one candidate executed per round.

\begin{table}[h]
\centering
\caption{Bayesian optimizer hyperparameters, held fixed wherever the
corresponding components are retained.}
\label{tab:app:bo-hyper}
\small
\begin{tabular}{lc}
\toprule
Hyperparameter & Value \\
\midrule
Trees per domain & 24 \\
Maximum tree depth & 4 \\
Feature subsampling fraction & 0.7 \\
Monte-Carlo draws $S$ & 64 \\
Warm-start threshold (feasible trials) & 5 \\
Warm-start weights (conf / novelty / risk) & 0.4 / 0.2 / 0.3 \\
Reference point & base model \\
Candidates executed per round & 1 \\
\bottomrule
\end{tabular}
\end{table}

\section{Instructions and Prompts}
\label{app:instructions}

This section presents the task instructions, skills, and prompt templates
for the reported experiments. The discovery-stage instruction is condensed
for presentation, with machine-specific operational rules omitted.
Placeholders indicate run-specific content inserted at call time.

\subsection{Discovery-Stage Task Instruction}
\label{app:instructions:agents}

The following condensed task instruction specifies the objective,
available support, deliverables, and rules for recipe-space design.

\begin{promptverb}
Construct a task-adapted post-training recipe space for the provided
base model and target benchmarks.

## Objective

Through data research and design, provide relevant, complementary,
and composable choices for subsequent search, including training
data sources, data processing methods, source combinations, and
cross-domain composition.

Submit a recipe space validated by the workflow and a valid initial
recipe selected from that space. The initial recipe demonstrates
executability; it does not specify the best eventual recipe.
This stage ends when the handoff is accepted.

## Autonomy & Exploration

- You may access the internet to research and compare multiple
  candidate datasets. Do not stop at the first loadable dataset.
- First study the benchmark documentation, implementation, and
  non-evaluation examples to understand the target capability,
  input and output requirements, scoring rules, and difficulty.
- Compare actual examples from candidate sources, considering
  task relevance, supervision quality, difficulty, and complementary
  coverage. Retain several meaningfully different training-source
  choices.
- Perform the necessary data filtering and field mapping, and
  validate the adapted data through the workflow.
- You may design Rewrite and/or Correction methods as needed.
  Specify a suitable source and generation prompts for each method.
  Use a separate source that excludes evaluation
  examples.
- Within the given data budget, design source combinations and
  cross-domain composition choices that support multiple
  executable recipes.
- You may iteratively revise sources, methods, and the recipe space
  based on workflow validation feedback.

## Information on the Setup

We provide benchmark descriptions, an evaluation interface, a
training-data budget, and a workflow for data adaptation, method
validation, and recipe-space handoff. Training-configuration choices
are supplied by the workflow; your design work focuses on data
sources, data processing, and data composition.

Use the skills in this order: benchmark, data, augmentation,
search_space, and handoff. They guide task analysis, source selection,
data-processing method design, recipe-space construction, and
handoff validation, respectively.

Create task.json, sources.json, search_space.json, runtime.json,
and initial_recipe.json in the provided formats. Initialize and
validate the run, then resolve issues identified by the workflow
until the handoff status is "accepted".

## Rules

1. There will be no user interaction. Operate autonomously.
2. Complete the task within the assigned time budget.
3. Do not use target benchmark evaluation questions or answers
   for training or training-data construction.
4. Do not modify the evaluation scripts or templates.
\end{promptverb}

\subsection{Discovery Skills}
\label{app:instructions:skills}

The initialization context $\mathcal{K}$ contains five skills that guide
benchmark analysis, data-source selection, data-processing method design,
recipe-space construction, and handoff validation. The excerpts below
retain their substantive instructions while omitting operational commands,
file-format specifications, and benchmark-specific output-format details.

\begin{promptverb}
=== SKILL: benchmark ===
Begin by studying every assigned benchmark. Use its implementation, scorer,
documentation, task format, and representative non-evaluation examples to
understand what successful training data should teach.

Create a concise working description of:

- the capability and knowledge being measured;
- the scorer-facing input and expected output;
- how correctness is determined;
- the approximate difficulty and content distribution;
- the kinds of examples likely to transfer to the benchmark.

Use this description when comparing data sources and designing data
processing methods.
Similarity includes capability, task form, and difficulty rather than storage
schema alone.

Target evaluation questions, answers, and transformations of them are
reserved for evaluation. Keep them out of training, data-generation inputs,
and generation prompts.
\end{promptverb}

\begin{promptverb}
=== SKILL: data ===
Research several candidate datasets for every benchmark. Begin with
established sources commonly used for the same or closely related capability,
then consider other sources when they add useful coverage. Compare actual
examples from the exact configuration and split; dataset names and schemas
alone are not evidence of suitability.

Judge candidates by:

- capability and content relevance;
- similarity of task form and difficulty;
- supervision quality and provenance;
- reproducibility and license;
- useful differences from the other source choices.

Select several genuinely useful base-training alternatives for each domain.
Quality and meaningful diversity matter more than filling the search space
with near-duplicates.

Prepare sources for three roles:

- Base training may use a broader range of relevant examples and should
  preserve the source's natural instruction format.
- Rewrite input should contain reliable existing supervision that the
  benchmark adapter can verify.
- Correction sources should be separate from training data and close
  enough to the benchmark to reveal relevant model failures.

Use deterministic preprocessing to select relevant examples and faithfully
convert existing fields into the required training format. A generally useful
dataset may be filtered when it contains examples outside the benchmark
capability - for example, filtering legal questions from a scientific
knowledge source. If the relevant subset cannot be selected reliably, choose
another dataset. Inspect the normalized output after conversion.
Verifier-backed roles need genuine source supervision such as an explicit
answer, label, or executable tests; never manufacture missing supervision
during adaptation.

Record dataset identity, revision, split, license, field mapping, and intended
role. Evaluation data and its derivatives are excluded from every role.
\end{promptverb}

\begin{promptverb}
=== SKILL: augmentation ===
Study the benchmark capability, task format, scorer, difficulty,
and representative non-evaluation examples.
Infer which variations could create useful training coverage and which
properties must remain valid.

Consider both depth and breadth when designing data generation, rather than
applying fixed transformations. Two methods are available:

- Rewrite evolves suitable source tasks into new tasks within the
  benchmark's capability neighborhood.
- Correction turns model failures on an independent, non-evaluation data
  source and the corresponding verifier feedback into corrected training
  examples.

Design Rewrite and/or Correction methods as needed. For each method, select
a suitable source and author generation prompts during recipe-space design.
The workflow invokes the method if it is selected during subsequent search;
small-scale validation during design checks whether it produces suitable
examples. The methods should offer substantive coverage rather than surface
paraphrases.

Every Rewrite uses two stages. The first prompt creates a new task input
by transforming the source task while preserving its relevance to the target
capability. The second prompt independently creates the required supervision
for the new input without seeing or copying the original answer. Correction
uses the latest trained model's errors on its independent source: its prompt
produces corrected supervision from the original task, wrong attempt, and
verifier feedback.

Validate each method through the workflow. Use example-level feedback to
improve the prompts, then validate the revised method again. Include only
accepted methods in the search space.

Design and validate each benchmark method independently. Use a validation
failure to repair the corresponding method and requirement.
Serialization, parser, token-limit, or stochastic generation failures are not
evidence that the task should be made easier. Do not propagate a workaround
from one benchmark to another.

Passing the adapter is necessary but not sufficient. Before registering a
method, inspect accepted examples for semantic correctness, benchmark-aligned
difficulty, reasoning depth, and diversity across the batch. Rewrite should
remain conditioned on the source task rather than collapse into repeated
free-form templates.
\end{promptverb}

\begin{promptverb}
=== SKILL: search_space ===
Construct a finite recipe space that offers meaningful data choices for
every benchmark.

For each domain, include:

- several accurate and complementary base-source choices;
- optionally, an accepted Rewrite method backed by a suitable source;
- optionally, an accepted Correction method backed by an independent,
  non-evaluation source;
- positive mixture-ratio choices.

Include cross-domain mixture modes that can represent different reasonable
allocations while keeping every objective covered. The resulting combinations
should provide multiple structurally different recipes rather than cosmetic
variations of one recipe.

The initial recipe is a valid executable example drawn from the choices;
it demonstrates that the space is runnable but does not define the eventual
best configuration.

\end{promptverb}

\begin{promptverb}
=== SKILL: handoff ===
Submit the data sources, data processing methods, recipe space, and a valid
initial recipe using the assigned base model and workflow-provided runtime
settings. Validate each included Rewrite or Correction method as described
in the augmentation skill.

The workflow checks source loading and format conversion, data-processing
validation results, and whether the initial recipe uses valid choices and
can be executed. Use the feedback to revise the corresponding sources,
methods, or choices and validate again.

The discovery task is complete when the workflow accepts the handoff.
\end{promptverb}

\subsection{Recipe-Proposer Prompt}
\label{app:instructions:proposer}

At every round of the optimization loop, the recipe proposer receives the
current recipe space and selectively retrieved trial evidence, as described
in Section~\ref{sec:recipe-optimizer}. The evidence includes strong and
lower-performing trials, domain-wise results, and choice-coverage statistics.
The proposer constructs multiple candidate recipes; trial selection is
handled separately. The template below omits field-level implementation
details. In the structured-memory ablation, the retrieved evidence is
replaced by the complete flat history of recipes and raw benchmark scores.

\begin{promptverb}
You are proposing candidate SFT recipes for multi-domain post-training.
Each recipe must be self-contained and use only current searchable choices.

## Searchable decision space
<Decision components and rules for using data generation>

## Current searchable choices
<Per-domain data sources, including previously generated datasets;
 available data processing methods; cross-domain mixture ratios and
 composition modes; SFT hyperparameters>

## Long-term search summary
<Reference-model scores, per-domain best results, statistics over explored
 choices, choice coverage, normalized Pareto front, and recently proposed
 but unevaluated candidates>

The best result in an individual domain does not necessarily identify the
best recipe across domains. Use these results and the Pareto front to propose
promising cross-domain combinations. Candidates listed as proposed-but-not-evaluated
have no measured outcome and must not be treated as failures. Use the
choice-coverage statistics to distinguish choices that have not been evaluated
from those associated with weak observed results.

## Candidate-set requirements
Treat all returned recipes as one candidate set and make them meaningfully
diverse from one another. The set must include exploitation of observed
evidence, recombination of promising choices, and exploration of
under-explored choices. Do not return minor variants that repeatedly
reuse the same per-domain data processing method.

## Historical trial feedback
<Selected historical trials, including Pareto-optimal or otherwise strong
 trials and lower-performing trials; each with its recipe, benchmark scores,
 domain-wise results, and aggregate score>

## Hard rules
- Select one available mixture ratio and one cross-domain composition mode.
  Specify data choices for every domain assigned a positive ratio.
- Select data sources and data processing methods only from the current
  choices for the corresponding domain.
- When selecting data generation, explicitly choose whether to use the
  generated data alone or combine them with existing sources. The workflow
  uses exactly the selected sources and generated data; it does not
  automatically add base sources.
- Infer source-mode preferences separately for each domain from historical
  empirical evidence; do not assume base-only, base+generated, or
  generated-only is universally superior.
- When evidence is incomplete or confounded, include controlled source-mode
  ablations that change one domain at a time.

## Required JSON schema
<Schema specifying each recipe's cross-domain composition, per-domain
 data sources and optional data generation, SFT hyperparameters, and
 expected effect, confidence, and risk>

## Example response format
<Two example recipes: one using interleaved base sources and one using
 generated data>

Return up to <N> candidate recipes in the required JSON format.
Do not include prose or markdown fences.
\end{promptverb}

\subsection{Baseline Task Prompt}
\label{app:instructions:baseline}

We use the original implementations of PostTrainBench and AutoTrainess
\citep{rank2026posttrainbench,yu2026autotrainess}, adapting only the task
objective and evaluation interface for joint post-training across six
benchmarks. Each baseline retains its original optimization procedure.
The condensed objective below follows the wording of the PostTrainBench
task prompt; \texttt{\{model\}} denotes the assigned base model.

\begin{promptverb}
## Objective

You should perform automated research and development to post-train
{model} to achieve the highest average score across the following
six benchmarks:

1. Mathematical reasoning: AIME 2025 and GSM8K.
2. Coding and tool use: BFCL v3 and HumanEval.
3. Science and health: GPQA Main and HealthBench-Easy.

You can evaluate the model via the evaluate.py script, which reports
individual benchmark scores and their mean within each domain.
The overall score is the mean of the three domain scores, equivalent
to the mean of all six benchmark scores.

Store the trained model with the highest overall score in final_model.
\end{promptverb}

\section{Additional Experimental Results}
\label{app:results}

We present the recipe spaces and generated data, the optimization
trajectories, and the remaining performance limitations of SRO.


\subsection{Optimization over a Predefined Recipe Space}
\label{app:predefined-space}

We instantiate a predefined-space baseline using SRO's execution
framework to optimize training configurations with a fixed training
dataset. The training pool consists of the first 5,000 examples
from the \texttt{openai/gsm8k} \texttt{main/train} split.
Source discovery, data mixture optimization, and data generation
are disabled.
The SFT search space contains eight LoRA configurations:
learning rate $\in\{10^{-4},2\times10^{-4}\}$,
training epochs $\in\{2,3\}$, and LoRA rank $\in\{16,64\}$.
Both model sizes use an effective batch size of 32, BF16 precision,
cosine learning-rate decay, a warmup ratio of 0.05, and gradient
checkpointing.

Candidates are sampled without replacement from the remaining
configurations using a fixed random seed, without an LLM proposer.
After cold-start initialization, a random-forest surrogate and
expected-improvement acquisition select the next trial using
GSM8K accuracy as the objective.
After five consecutive SFT trials without improvement in the
best observed score, the procedure enters a GRPO phase initialized
from the best SFT checkpoint.
Both runs evaluate six of the eight SFT configurations before
this transition.
The final model is evaluated on all 1,319 GSM8K test examples.

\begin{table}[t]
\centering
\caption{GSM8K accuracy (\%) under optimization over a predefined
recipe space with a fixed training dataset. Single-domain SRO
results with DeepSeek-V4-Flash from Table~1 are reproduced
for reference.}
\label{tab:predefined-space}
\small
\begin{tabular}{lrrrrr}
\toprule
Base model & Before training & Best SFT & GRPO & Final & SRO (Ours) \\
\midrule
Qwen3-1.7B-Base & 10.3 & 55.0 & 55.7 & 55.7 & 68.2 \\
Qwen3-4B-Base   & 43.0 & 74.8 & 74.5 & 74.8 & 87.6 \\
\bottomrule
\end{tabular}
\end{table}

Table~\ref{tab:predefined-space} summarizes the results.
Both models obtain their best SFT result with a learning rate of
$2\times10^{-4}$, three training epochs, and LoRA rank 64.
GRPO further improves Qwen3-1.7B to 55.7, whereas Qwen3-4B retains
its best SFT model at 74.8.
This comparison measures performance when task-specific training
data are supplied in advance and search is restricted to training
configuration.

\subsection{Agent-Constructed Recipe Spaces}
\label{app:discovery}

Each base-model and agent-LLM configuration independently constructs
its data-source choices, data processing methods, and recipe space.

\paragraph{The constructed spaces contain complementary data sources and processing choices.}
\label{app:discovery:sources}

Table~\ref{tab:app:discovery-ds} presents the data-source choices from the
DeepSeek-V4-Flash discovery run for Qwen3-1.7B. For each benchmark,
the agent selected base training sources, a Rewrite input source, and a
separate Correction source; it also designed one Rewrite and one Correction
method (Table~\ref{tab:app:discovery-methods}).

\begin{table}[h]
\centering
\caption{Discovery-selected sources for DeepSeek-V4-Flash with Qwen3-1.7B.
Correction uses separate source data that exclude evaluation examples.}
\label{tab:app:discovery-ds}
\small
\setlength{\tabcolsep}{3.5pt}
\begin{tabularx}{\linewidth}{@{}lXXX@{}}
\toprule
Benchmark & Base training sources & Rewrite source & Correction source \\
\midrule
AIME 2025 & AIME 1983--2024; Hendrycks MATH & AIME 1983--2024 & \sid{aime_diag} \\
GSM8K & GSM8K train; MetaMathQA (GSM subset) & GSM8K train & \sid{gsm8k_diag} \\
HumanEval & MBPP; MBPP+ (EvalPlus) & MBPP & \sid{mbpp_diag} \\
BFCL & Hermes function-calling; Glaive v2 & Hermes & \sid{bfcl_hermes_diag} \\
GPQA Main & MMLU (science subset); ARC & MMLU (science subset) & \sid{arc_diag} \\
HealthBench & MedAlpaca flashcards; MedAlpaca WikiDoc & MedAlpaca flashcards & medical-o1-reasoning \\
\bottomrule
\end{tabularx}
\end{table}

\begin{table}[h]
\centering
\caption{Agent-designed data-processing methods in the DeepSeek-V4-Flash
run for Qwen3-1.7B. Rewrite uses separate task-generation and
answer-generation prompts; Correction generates supervision from model
errors on independent source data.}
\label{tab:app:discovery-methods}
\small
\setlength{\tabcolsep}{3.5pt}
\begin{tabularx}{\linewidth}{@{}llX@{}}
\toprule
Benchmark & Method & Description \\
\midrule
\multirow{2}{*}{AIME} & Rewrite & Evolve AIME-adjacent competition problems into new integer-answer (0--999) problems with independent full reasoning. \\
 & Correction & Turn wrong AIME-style answers on independent source data into corrected full-reasoning integer-answer solutions. \\
\midrule
\multirow{2}{*}{GSM8K} & Rewrite & Evolve GSM8K-style word problems into new multi-step arithmetic problems with independent solutions. \\
 & Correction & Turn wrong GSM8K-style answers on independent source data into corrected full-reasoning numeric solutions. \\
\midrule
\multirow{2}{*}{HumanEval} & Rewrite & Evolve MBPP-style tasks into new function-writing specs with independent solutions and executable tests. \\
 & Correction & Repair failing MBPP-style solutions on independent source data into correct, test-passing implementations. \\
\midrule
\multirow{2}{*}{BFCL} & Rewrite & Rewrite function-calling conversations into new grounded requests with a single schema-valid tool call. \\
 & Correction & Correct wrong BFCL-style tool calls into schema-valid single calls that fulfill the request. \\
\midrule
\multirow{2}{*}{GPQA} & Rewrite & Evolve MMLU-style science questions into new four-choice questions with independent reasoned answers. \\
 & Correction & Repair wrong science multiple-choice answers on independent source data into reasoned, correctly-labeled solutions. \\
\midrule
\multirow{2}{*}{HealthBench} & Rewrite & Evolve medical Q\&A conversations into new clinical scenarios with complete, safe, uncertainty-aware answers. \\
 & Correction & Repair wrong or unsafe health answers on independent source data into complete, safe, uncertainty-aware, actionable responses. \\
\bottomrule
\end{tabularx}
\end{table}

\paragraph{The selected sources vary across models and agent LLMs.}
\label{app:discovery:comparison}

For Qwen3-4B, DeepSeek-V4-Flash independently constructed a different recipe
space. Its sources include AIME gneubig, OpenR1, and MATH for AIME;
GSM8K, MetaMath, and OpenMathInstruct for GSM8K; MBPP full and sanitized
variants for HumanEval; Hermes and Glaive for BFCL; MMLU STEM and
professional subsets and MMLU-Pro for GPQA; and MedQA, MedMCQA, and
MMLU medical subsets for HealthBench. Both DeepSeek configurations include
Rewrite and Correction choices for each benchmark.

For Qwen3-1.7B, GPT-5.4 performed independent source research
under the same skills and constraints and selected partially different
sources: for mathematics, NuminaMath (general subset) and a
competition-training pool replace the AIME/MATH archive; for GSM8K, Orca-Math
word problems are added; for GPQA, SciQ and PubMedQA replace MMLU/ARC; for
HealthBench, ChatDoctor replaces MedAlpaca, with MedMCQA and medical-o1
serving as Correction sources. This run includes one Rewrite and one
Correction method per benchmark.

\paragraph{Examples illustrate the resulting training data.}
\label{app:cases:examples}

We present lightly formatted excerpts from three accepted generated rows
in the DeepSeek-V4-Flash--Qwen3-1.7B SRO run: two Rewrite examples
(GSM8K and BFCL) and one Correction example (GPQA). Formatting is condensed
for readability, while the task content, generated supervision, and
verification outcomes are preserved. The BFCL tool call is summarized,
and the GPQA rationale is excerpted.

The two Rewrite rows are recorded as \sid{teacher_labeled_and_verified}:
their new inputs and supervision were generated separately and accepted
by the corresponding benchmark adapters. The GPQA Correction row is
recorded as \sid{verifier_corrected}: it was generated from an actual
checkpoint failure on an independent diagnostic source and checked against
that source's reference answer. All three examples originate from training
or independent diagnostic sources, not benchmark evaluation data.

\begin{promptbox}{Rewrite example: arithmetic reasoning}
\small
\textbf{Training prompt:} Solve the following math problem
step by step. The last line of your response should be of the form
``ANSWER: \$ANSWER'' (without quotes) where \$ANSWER is the answer to the
problem.\par
``Maya is organizing a school fundraiser. She buys 5 boxes of cookies, each
containing 12 packs. Each pack has 8 cookies. She sells 3/4 of the total
cookies at a bake sale. Later, she donates half of the remaining cookies to
a local shelter. How many cookies does Maya have left after donating?''\par
\medskip
\textbf{Assistant (independently generated supervision):} First, calculate
the total number of packs: 5 boxes $\times$ 12 packs per box = 60 packs. Then
calculate total cookies: 60 packs $\times$ 8 cookies per pack = 480 cookies.
She sells 3/4 of the total cookies: 3/4 $\times$ 480 = 360 cookies sold.
Remaining after sale: 480 $-$ 360 = 120 cookies. She donates half of the
remaining: 1/2 $\times$ 120 = 60 cookies donated. Cookies left after
donating: 120 $-$ 60 = 60.\par
ANSWER: 60
\par\medskip
\textbf{Source and verification:}
\sid{math_gsm8k_rewrite_v1} uses the training source
\sid{math_gsm8k_train}. Of 500 generated rows, 481 passed validation
(96.2\%).
\end{promptbox}

\begin{promptbox}{Rewrite example: tool use}
\small
\textbf{User:} ``Can you please program my office thermostat (unit ID:
office-thermo-7) so that on Monday through Friday it warms up to 70 degrees
at 6:30 AM and cools down to 60 degrees at 10:00 PM, and on Saturday and
Sunday it stays at 72 degrees starting at 9:00 AM and drops to 64 degrees at
11:30 PM?''\par
\medskip
\textbf{Assistant (tool-call summary):} a single call
to \sid{set_thermostat_schedule} with \sid{thermostat_id} set to
\sid{"office-thermo-7"} and a \sid{schedule} array of fourteen
(\sid{day}, \sid{temperature}, \sid{time}) entries, e.g.\
(Monday, 70, 06:30), (Monday, 60, 22:00), \ldots, (Sunday, 72, 09:00),
(Sunday, 64, 23:30).\par
\medskip
\textbf{Tools supplied to the model:} one function,
\sid{set_thermostat_schedule}, with a typed JSON schema over
\sid{thermostat_id} and the \sid{schedule} array. The rewrite stage created
a new request grounded in this tool set; the answer stage produced the
single schema-valid call that fulfills it.
\par\medskip
\textbf{Source and verification:}
\sid{code_bfcl_rewrite_v1} uses the external training source
\sid{code_bfcl_hermes}. Of 350 generated rows, 331 passed the schema
verifier (94.57\%).
\end{promptbox}

\begin{promptbox}{Correction example: science reasoning}
\small
\textbf{Training prompt:} Answer the following multiple choice
question. Reason step by step. The last line of your response should be of
the form ``ANSWER: \$LETTER'' where \$LETTER is one of A, B, C, or D.\par
``An astronaut drops a 1.0 kg object and a 5.0 kg object on the Moon. Both
objects fall a total distance of 2.0 m vertically. Which of the following
best describes the objects after they have fallen a distance of 1.0 m?''\par
A. They have each lost kinetic energy. \quad B. They have each gained the
same amount of potential energy. \quad C. They have each lost the same amount
of potential energy. \quad D. They have each gained one-half of their maximum
kinetic energy.\par
\medskip
\textbf{Original checkpoint answer:} \sid{ANSWER: C}.
\textbf{Reference answer:} D.\par
\medskip
\textbf{Assistant (corrected supervision):} The gravitational force on the
Moon is the same for both objects in terms of acceleration, but the potential
energy change is $mgh$, so the two objects do not lose the same amount of
potential energy. The maximum kinetic energy is reached just before hitting
the ground after falling 2.0 m; after falling 1.0 m they have gained half
their maximum kinetic energy, regardless of mass. [...]\par
ANSWER: D\par
\medskip
\textbf{Source and verification:}
\sid{science_gpqa_correction_v1} uses the independent ARC source
\sid{science_gpqa_arc_diag}, not the GPQA evaluation set. The SFT checkpoint
from trial 9 answered 97 of 500 diagnostic examples incorrectly.
Correction produced verified responses for 96 of these 97 examples
(98.97\%); the displayed example was corrected from C to D.
\end{promptbox}

\subsection{Recipe Optimization Trajectories}
\label{app:dynamics}

\paragraph{The ablation trajectories complement the final-score comparisons.}
\label{app:dynamics:curves}

\begin{figure}[h]
\centering
\begin{minipage}[t]{0.48\textwidth}
\centering
\includegraphics[width=\linewidth]{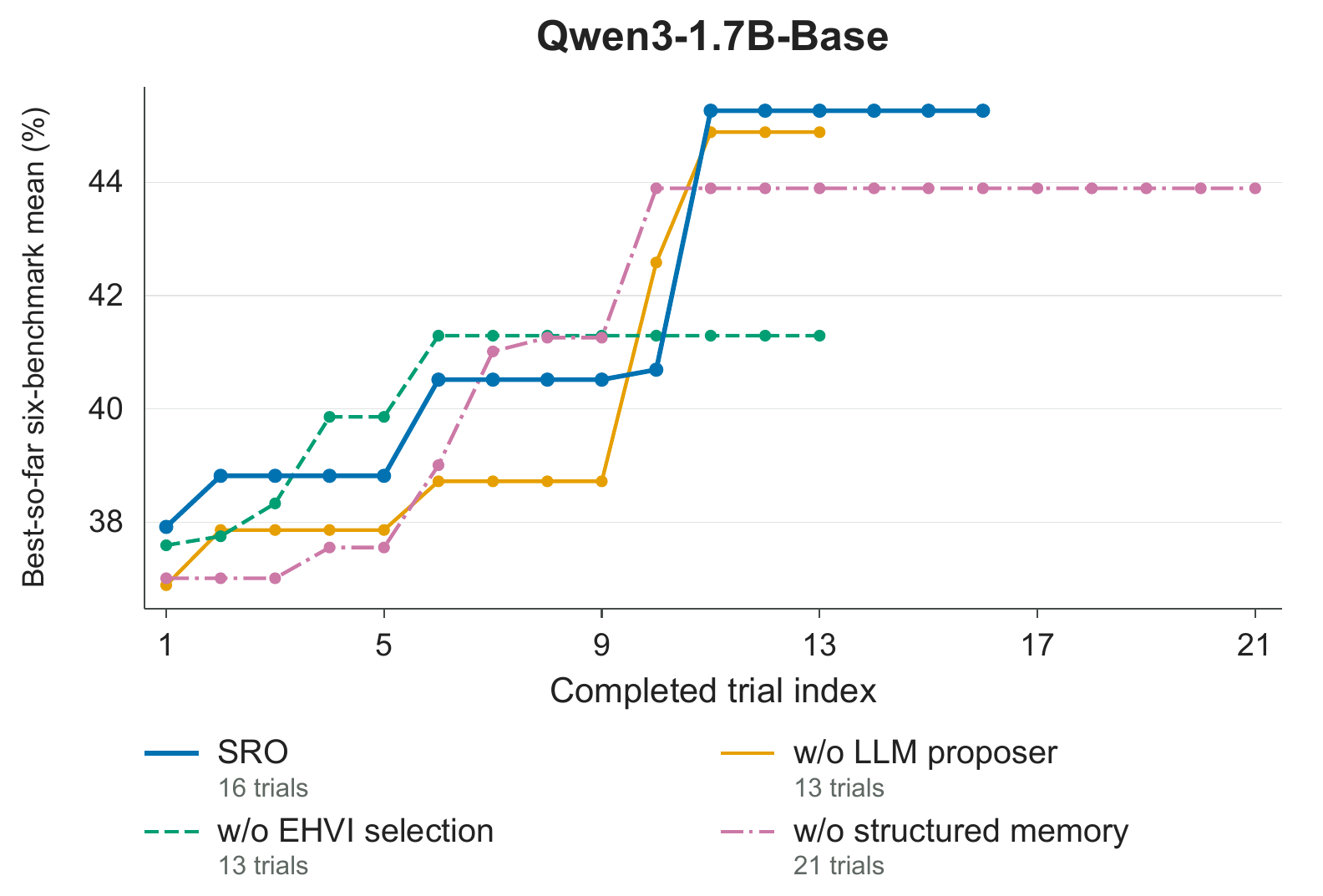}
\end{minipage}\hfill
\begin{minipage}[t]{0.48\textwidth}
\centering
\includegraphics[width=\linewidth]{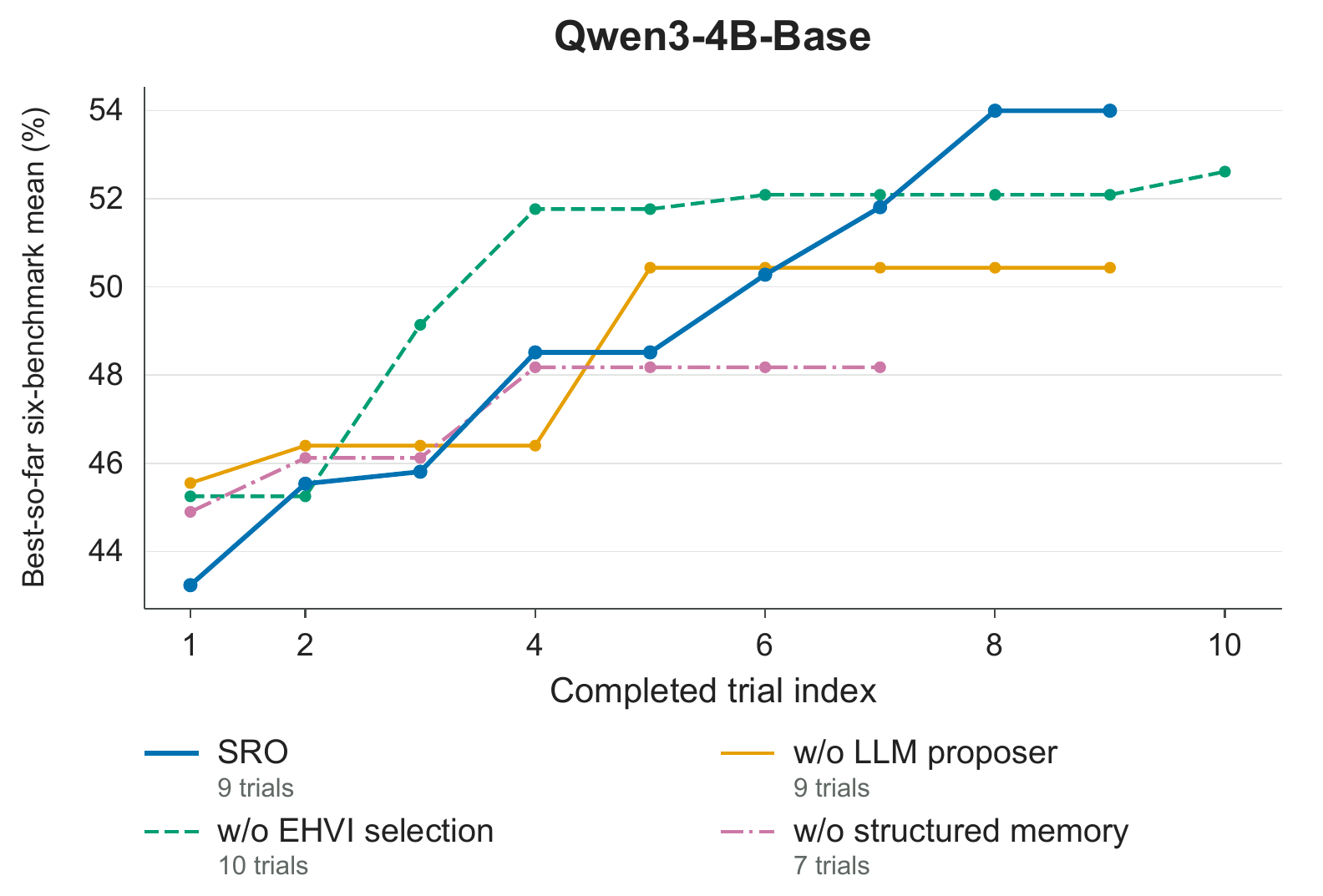}
\end{minipage}
\caption{Best-so-far search-time six-benchmark mean versus completed
trial index for SRO and three optimizer ablations.
Left: Qwen3-1.7B-Base. Right: Qwen3-4B-Base.
All runs use DeepSeek-V4-Flash and the same compute budget.
Trial counts differ because recipe execution costs vary.}
\label{fig:app:curves}
\end{figure}

Figure~\ref{fig:app:curves} plots the best-so-far search-time
six-benchmark mean against completed trials for SRO and three optimizer
ablations, all using DeepSeek-V4-Flash. For each target model, the full
system and these ablations use the same validated initial recipe space
and compute budget. The two target models have independently constructed
spaces. The \textit{w/o structured recipe search} variant is not included
in this trial-indexed comparison because it does not use the same
recipe-execution workflow.

The trajectories show when gains are obtained and whether subsequent
trials improve the incumbent. On both target models, full SRO achieves
a higher final search-time score than the three ablations, providing
additional evidence for the contributions of LLM candidate construction,
EHVI selection, and structured memory. Trial-indexed curves describe
progress per completed experiment; the number of experiments completed
within the time budget also depends on data generation, training, and
evaluation costs.

\paragraph{Search progress varies across runs.}
\label{app:results:trials}

Table~\ref{tab:app:search-summary} summarizes the completed trial counts,
selected trials, and corresponding search-time and final evaluation
scores for the four multi-domain runs. In several runs, the best recipe
is found before search terminates, and subsequent trials do not further
improve the best search-time aggregate score.

\begin{table}[h]
\centering
\caption{Search summary of the four SRO multi-domain runs.
The selected trial maximizes the search-time six-benchmark mean.
Final scores are obtained by evaluating the selected checkpoint on the
complete benchmark sets after the time budget.
Evaluation settings are detailed in Appendix~\ref{app:setup:benchmarks}.
All scores are reported on a 0--100 scale.}
\label{tab:app:search-summary}
\small
\setlength{\tabcolsep}{4pt}
\begin{tabular}{llrrrr}
\toprule
Agent LLM & Base model & Trials & Selected trial & Search-time & Final \\
\midrule
DeepSeek-V4-Flash & Qwen3-1.7B & 16 & 11 & 45.3 & 45.3 \\
DeepSeek-V4-Flash & Qwen3-4B   &  9 &  8 & 54.0 & 53.8 \\
GPT-5.4          & Qwen3-1.7B & 17 &  7 & 40.9 & 41.1 \\
GPT-5.4          & Qwen3-4B   &  8 &  8 & 48.3 & 47.5 \\
\bottomrule
\end{tabular}
\end{table}

\paragraph{Higher aggregate scores can coexist with lower individual
benchmark scores.}
\label{app:dynamics:tradeoffs}

During the DeepSeek-V4-Flash run on Qwen3-1.7B, some recipes achieve
strong GSM8K scores while substantially degrading GPQA performance;
the selected recipe also scores below the run's GPQA peak.
On Qwen3-4B, an increase in the aggregate score similarly coincides
with a decrease in AIME performance. These observations illustrate
the cross-domain trade-offs in joint post-training: recipe quality
must be assessed across the target domains rather than from any
individual benchmark alone. Because recipes can differ in several
data and training choices, these observations do not isolate the
effect of any single component.
\subsection{Limitations and Failure Cases}
\label{app:cases}
\label{app:cases:failures}

\paragraph{Competition mathematics remains difficult.}
Across the four multi-domain runs, the final selected models achieve
AIME scores of 0.0--6.7, substantially below the instruction-tuned
reference scores of 26.7 for Qwen3-1.7B and 53.3 for Qwen3-4B.
With only 30 evaluation problems, each additional correct answer
changes the score by about 3.3 points. These results show that a high
aggregate score does not imply strong performance on every capability.

\paragraph{Recipe validity does not guarantee capability preservation.}
During the DeepSeek-V4-Flash run on Qwen3-1.7B, some candidate
recipes produce severe regressions in BFCL performance.
Such failures show that recipes assembled from admissible choices
can still perform poorly on an individual benchmark.
Recipe validation checks executability but does not guarantee
preservation of every capability. The observed scores alone do not
determine whether these regressions arise from data composition,
training settings, or output-format failures.

\section{Contamination Safeguards and Run Hygiene}
\label{app:contamination}

Because automated post-training systems have been shown to engage in reward
hacking when given the opportunity \citep{rank2026posttrainbench}, all
reported runs enforce the following safeguards. These apply identically to
SRO and both baselines.

\paragraph{Task-level rules.} Every run receives the contamination rules
reproduced in Appendix~\ref{app:instructions:baseline}: benchmark questions,
answers, solutions, tool-call targets, rubrics, and transformations of them
are reserved for evaluation and must not appear in training sources,
Rewrite inputs, diagnostic sources, generated-data prompts, or examples.

\paragraph{Structural guards in SRO.} Beyond the task-level rules, the SRO
workflow enforces contamination safety structurally, since the search space
is machine-checked rather than free-form code:
(i)~every declared source passes a benchmark adapter that canonicalizes rows
and validates required supervision (explicit numeric targets for math,
four choices plus an A--D gold label for GPQA, executable tests plus
reference solutions for HumanEval, schema-valid tool calls for BFCL, complete
multi-turn structure for HealthBench), and rejected sources cannot appear in
recipe choices;
(ii)~HealthBench rows containing official \sid{rubrics},
\sid{ideal_completions_data}, or \sid{canary} fields are rejected outright;
(iii)~Rewrite supervision is generated by an answer stage that never sees the
source target, and Correction supervision is verified against the diagnostic
source's own oracle rather than against any benchmark artifact;
(iv)~diagnostic sources are held out from training-data roles and used only
to surface model failures for Correction;
(v)~the handoff records a content digest of every source and method contract,
so the exact training data of every reported run is reproducible and
auditable after the fact.

\paragraph{Model substitution.} Only the assigned base model may be
fine-tuned; checkpoints must derive from the base model. SRO satisfies this
by construction (the workflow trains only from the assigned base checkpoint),
and the baselines receive the same rule in their task prompt. Final
evaluation loads each submitted \sid{final_model} through the same
benchmark entrypoints used during the run, with evaluation scripts and
templates immutable from the agent's perspective.

\end{document}